\pdfoutput=1
\documentclass{article}
\usepackage[main, final]{neurips_2026}

\usepackage[utf8]{inputenc}
\usepackage[T1]{fontenc}
\usepackage{hyperref}
\usepackage{url}
\usepackage{booktabs}
\usepackage{amsfonts}
\usepackage{amsmath}
\usepackage{amssymb}
\usepackage{amsthm}
\usepackage{nicefrac}
\usepackage{microtype}
\usepackage{xcolor}
\usepackage{graphicx}
\usepackage{multirow}
\usepackage{wrapfig}
\usepackage{pifont}
\usepackage{tcolorbox}
\tcbuselibrary{skins}
\usepackage{caption}
\usepackage{graphicx}     %
\usepackage{subcaption}   %
\usepackage{amssymb}      %
\usepackage[table]{xcolor} %
\usepackage{cleveref}
\usepackage{amsmath}  %
\usepackage{amssymb}  %
\usepackage{amsfonts} %
\usepackage{enumitem}
\usepackage{rotating}

\newtheorem{theorem}{Theorem}

\newtheorem{example}{Example}
\newtheorem{corollary}{Corollary}

\newcommand{\bad}{\ding{55}}

\tcolorboxenvironment{theorem}{
    boxrule=0pt, frame hidden, left=6pt, right=6pt, top=4pt, bottom=4pt,
    before skip=12pt, after skip=12pt, colback=gray!10
}
\tcolorboxenvironment{lemma}{
    boxrule=0pt, frame hidden, left=6pt, right=6pt, top=4pt, bottom=4pt,
    before skip=12pt, after skip=12pt, colback=gray!10
}
\tcolorboxenvironment{corollary}{
    boxrule=0pt, frame hidden, left=6pt, right=6pt, top=4pt, bottom=4pt,
    before skip=12pt, after skip=12pt, colback=gray!10
}
\tcolorboxenvironment{example}{
    boxrule=0pt, frame hidden, left=6pt, right=6pt, top=4pt, bottom=4pt,
    before skip=12pt, after skip=12pt, colback=gray!10
}

\newcommand{\R}{\mathbb{R}}
\newcommand{\C}{\mathbb{C}}

\newcommand{\dec}{\mathrm{dec}}

\newif\ifshowchanges
\showchangesfalse
\ifshowchanges
  
  \newcommand{\why}[1]{{\color{blue}\footnotesize\,[#1]}}
  \newenvironment{changed}{\color{red}}{}
  \DeclareCaptionFont{chgred}{\color{red}}
  \newenvironment{changedfloat}{\let\chgcentering\centering
    \def\centering{\chgcentering\color{red}}\captionsetup{font+=chgred}}{}
\else
  
  \newcommand{\why}[1]{\ifhmode\unskip\fi}

\fi
\newif\ifshowdeltarule
\ifdefined\markdeltarule\showdeltaruletrue\fi
\ifshowdeltarule
  
  \newenvironment{drchanged}{\color{blue}}{}
  \DeclareCaptionFont{drblue}{\color{blue}}
  \newenvironment{drfloat}{\let\drcentering\centering
    \def\centering{\drcentering\color{blue}}\captionsetup{font+=drblue}}{}
\else

\fi
\newcommand{\cs}[1]{c^{\star}_{#1}}

\title{Rethinking State Tracking in Recurrent Models Through Error Control Dynamics}

\author{Jiwan Chung, Heechan Choi, Seon Joo Kim \\
Yonsei University\\
\texttt{jiwan.chung.research@gmail.com}\\
\url{https://jiwanchung.github.io/rethinking-state-tracking/}}

\hypersetup{pdftitle={Rethinking State Tracking in Recurrent Models Through Error Control Dynamics}, pdfauthor={Jiwan Chung, Heechan Choi, Seon Joo Kim}}

\begin{document}

\maketitle

\begin{abstract}
The theory of state tracking in recurrent architectures has predominantly focused on expressive capacity: whether a fixed architecture can theoretically realize a set of symbolic transition rules. We argue that equally important is error control, the dynamics governing hidden-state drift along the directions that distinguish symbolic states. We prove that affine recurrent networks, a class of models encompassing State-Space Models and Linear Attention, cannot correct errors along state-separating subspaces once they preserve state representations. Consequently, practical affine trackers do not learn robust state tracking; rather, they learn finite-horizon solutions governed by accumulated state-relevant error. We characterize the mechanics of this failure, showing that tracking remains readable only while the accumulating within-class spread remains small relative to the between-class separation. We demonstrate empirically on group state-tracking tasks that this breakdown is predictable: tracking collapses when the distinguishability ratio crosses the readability threshold. Across trained models, the point of this crossing predicts the horizon at which downstream accuracy fails. These results establish that robust state tracking is determined not only by an architecture's theoretical expressivity but crucially by its error control.
\end{abstract}

\section{Introduction}
\label{sec:intro}

The theory of state tracking in recurrent architectures has been predominantly a theory of expressivity: which symbolic transition rules can a fixed architecture in principle realize \citep{merrill2024illusion,sarrof2024expressive,grazzi2024unlocking,karuvally2025aussm,shakerinava2026expressive}. We argue that equally important is \emph{error control}, the dynamics governing hidden-state drift along the directions that distinguish symbolic states. We prove that affine recurrences, a class that includes State-Space Models (SSMs) \citep{gu2022efficiently} as well as Linear Attention \citep{katharopoulos2020transformers}, cannot correct hidden-state drift along state-separating subspaces once they preserve state representations exactly.

In practice the two requirements diverge. Recent literature documents this gap: input-dependent complex-diagonal SSMs sufficient for $S_3$ at depth two fail to learn solutions that generalize beyond the training length \citep{shakerinava2026expressive}, and diagonal selective SSM variants can fit regular-language emulation at training lengths while collapsing under length extrapolation \citep{terzic2025expressiveness}. The same pattern surfaces within an architecture's own claimed task scope: AUSSM, provably sufficient for Abelian groups via unit-modulus rotations \citep{karuvally2025aussm,shakerinava2026expressive}, tracks $C_2$ and $C_6$ unevenly in our experiments. Across recurrent architectures developed for long-context sequence modeling \citep{gu2024mamba,lahoti2026mamba3,karuvally2025aussm}, expressive capacity does not predict state-tracking robustness.

In this work, we study error control as the missing axis in recurrent state tracking. We first show that affine recurrent models cannot correct symbolic-state drift once they preserve state representations (\Cref{subsec:theory_impossibility}). State-dependent return maps escape this obstruction and can selectively contract symbolic-subspace drift; we verify which canonical activations realize this correction (\Cref{ax:nonlinearity_jacobians}). We then characterize the finite horizons that affine trackers sustain without state-dependent correction. Their failures are governed by accumulated state-relevant error: tracking remains readable while within-class spread remains small relative to between-class separation, and breaks down once this ratio crosses the readability threshold (\Cref{subsec:theory_error_budget}).

We evaluate this account on a set of group state-tracking tasks. Performance exhibits a systematic separation: state-dependent models maintain tracking over the longest tested horizons, whereas affine models lose accuracy at different horizons (\Cref{subsec:exp_downstream}). This variation is central to our analysis: affine trackers are not distinguished only by whether they fail, but by how long they sustain tracking under repeated recurrence.

Our diagnostics give a consistent error-dynamics explanation.
Perturbation recovery shows that state-dependent models selectively contract injected hidden-state errors, whereas affine models do not (\Cref{subsec:exp_perturbation}).
The absence of selective contraction need not cause immediate failure: the distinguishability ratio $q(t)=R(t)/M(t)$ tracks how affine models gradually exhaust a finite horizon as within-class spread approaches between-class separation (\Cref{subsec:exp_codebook_collapse}), and subspace decomposition localizes this spread along the symbolic subspace $\mathcal{U}$, where affine return maps cannot contract errors (\Cref{subsec:exp_subspace_decomp}).
The point at which $q(t)$ first crosses the readability threshold, denoted $T_{\mathrm{cross}}$, quantitatively predicts the downstream max-passing length across affine sweeps on $S_3$ (\Cref{fig:corr}), confirming the finite-horizon mechanism of \Cref{cor:error_budget_modes}.

Together, these results establish that robust state tracking is determined not only by an architecture's theoretical expressivity but crucially by its error control dynamics.

\section{Background}
\label{sec:background}

\subsection{Recurrent models}
\label{subsec:background_models}

We provide a taxonomy of recurrent models explored in this work, from SSMs to general RNNs as shown in~\Cref{tab:model_properties}. We begin by introducing a common recursive form.

\begin{tcolorbox}[
    boxrule=0pt, 
    frame hidden,
    left=6pt, 
    right=6pt, 
    top=4pt, 
    bottom=4pt,
    before skip=12pt, %
    after skip=12pt,  %
    colback=gray!10 %
]
\textbf{Definition 1} (Recursive layer).
A \(d\)-dimensional recursive layer is a parameterized function that takes as input a sequence \(x_t \in \mathcal{X}\) and produces outputs \(y_t \in \mathcal{Y}\) via the recurrence
\begin{align}
h_t &= \phi\!\Big( g(h_{t-1},x_t)\odot\big(A(x_t)h_{t-1}\big) + b(x_t) \Big), \label{eq:canonical_state}\\
y_t &= \dec(h_t,x_t), \label{eq:canonical_output}
\end{align}
where \(\mathbb F\in\{\mathbb R,\mathbb C\}\), \(h_t \in \mathbb{F}^d\) is the latent state, \(A(x_t)\in\mathbb{F}^{d\times d}\) is the state-transition matrix, \(b(x_t)\in\mathbb{F}^d\) is the input-dependent injection term, \(g:\mathbb{F}^d\times\mathcal{X}\to\mathbb{F}^d\) is a state-dependent modulation, \(\phi:\mathbb{F}^d\to\mathbb{F}^d\) is an optional output nonlinearity, and \(\dec:\mathbb{F}^d\times\mathcal{X}\to\mathcal{Y}\) is a decoder.
\label{def:recursive_layer}
\end{tcolorbox}

Equation~\eqref{eq:canonical_state} isolates four conceptually distinct ingredients: state transition \(A(x_t)h_{t-1}\), input injection \(b(x_t)\), state-dependent modulation \(g(h_{t-1},x_t)\), and output activation function \(\phi\). Different model classes arise by constraining or removing these ingredients.

SSMs such as S4~\citep{gu2022efficiently} lie in the affine-in-state regime with \(g \equiv \mathbf{1}\) and \(\phi(z)=z\), using structured transition operators \(A\). Mamba~\citep{gu2024mamba} makes transition parameters input-adaptive, i.e., \(A=A(x_t)\) and \(b=b(x_t)\). Mamba-3~\citep{lahoti2026mamba3} and AUSSM~\citep{karuvally2025aussm} further increase the expressivity of this family through complex-valued state-space dynamics. More general linear recurrent models allow non-diagonal or matrix-valued transitions \(A(x_t)\), as in DeltaNet~\citep{yang2024deltanet} and DeltaProduct~\citep{siems2025deltaproduct}. Conventional RNNs~\citep{elman1990finding} introduce a nonlinear activation \(\phi\), while gated models~\citep{hochreiter1997lstm,cho2014learning} introduce state-dependent gating, which we capture conceptually through the multiplicative modulation \(g(h_{t-1},x_t)\). Refer to \Cref{ax:model} for details.

\begin{table*}[t]
\centering
\small
\setlength{\tabcolsep}{6pt}
\renewcommand{\arraystretch}{1.15}
\begin{tabular}{llll}
\toprule
\textbf{Dynamics} & \textbf{Model} & \textbf{Transition $A$} & \textbf{Field} \\
\midrule

\multirow{7}{*}{Affine}
& Mamba~\citep{gu2024mamba}                       & diagonal                    & real    \\
& Mamba-3~\citep{lahoti2026mamba3}                & diagonal                    & complex \\
& AUSSM~\citep{karuvally2025aussm}                & diagonal (unitary)          & complex \\
& Simple AUSSM~\citep{shakerinava2026expressive}  & diagonal (unitary)          & complex \\
& Negative Mamba~\citep{grazzi2024unlocking}  & diagonal (signed)           & real    \\
& Linear RNN                                       & dense                       & real    \\
& Token-gated RNN                                  & dense, input-gated          & real    \\

\midrule

\multirow{2}{*}{State-dependent}
& tanh RNN~\citep{elman1990finding}                & dense                       & real    \\
& State-gated RNN                                  & dense, state-gated          & real    \\

\bottomrule
\end{tabular}
\caption{
\textbf{Recurrent models categorized by properties of the state-transition matrix $A$.} A model is \emph{affine} when the state Jacobian $\partial h_t/\partial h_{t-1}$ does not depend on $h_{t-1}$, and \emph{state-dependent} otherwise. \emph{Transition $A$} describes the structure of the linear part of the recurrence; \emph{Field} indicates whether $A$ is real- or complex-valued. Full operator definitions in \Cref{ax:model}.
}
\label{tab:model_properties}
\end{table*}

\subsection{State \texorpdfstring{tracking and groups}{tracking and groups}}
\label{subsec:background_task}

\emph{State tracking} is the problem of maintaining a latent representation of
a symbolic state that evolves under an input sequence. Let \(G\) be a finite
state space and let \(\mathcal T:G\times\mathcal X\to G\) be a transition
rule. Given \(g_0\in G\) and inputs \(x_1,\ldots,x_L\), the symbolic trajectory
is
\[
g_t=\mathcal T(g_{t-1},x_t),\qquad t=1,\ldots,L .
\]
A model receives the sequence online and must maintain enough information in
its hidden state \(h_t\) to recover \(g_t\) at each step.

A convenient class of state-tracking tasks is given by finite groups. A
\emph{group} is a set \(G\) with an associative binary operation, an identity
element, and inverses. When inputs \(x_t\) are drawn from generators
\(\Sigma\subset G\), the transition is group multiplication,
\(g_t=g_{t-1}\cdot x_t\). The target is the running product
\(y_t=x_1\cdot x_2\cdots x_t\) after each input. See \citet{rotman2012introduction} for more details. We evaluate models on several
groups that vary in compositional structure:

\paragraph{Parity and cyclic groups (\(C_k\)).}
Cyclic groups represent modular counting and are generated by a single element;
\(C_2\) is the parity task. All cyclic groups are Abelian, so reordering input
group elements does not change the final product.

\paragraph{Symmetric groups (\(S_k\)).}
The symmetric group \(S_k\) consists of all permutations of \(k\) elements. For
\(k\ge 3\), \(S_k\) is non-Abelian, so input order changes the resulting state.
Thus \(S_3\) is the smallest symmetric group where order-sensitive composition
is unavoidable.

\begin{example}[\(S_3\)]
Let \(S_3\) be the set of permutations of \(\{1,2,3\}\), and let the input tokens be generators \((12)\) and \((23)\). Starting from the identity \(g_0=e\), the sequence \((12),(23),(12)\) yields
\[
g_1=(12),\qquad g_2=(123),\qquad g_3=(13).
\]
The task is to output the running product after each token.
\end{example}

\section{Error control in state tracking}
\label{sec:theory}
Prior work often studies recurrent architectures through \emph{expressivity}:
whether a continuous-state model can realize a symbolic transition rule. For
long-horizon state tracking, however, exact realization on clean trajectories
is not enough. A robust tracker must also correct hidden-state perturbations
that move it toward an incorrect symbolic state.

\subsection{Exact affine tracking cannot correct state error}
\label{subsec:theory_impossibility}

Let \(G\) be a finite symbolic state space, with each \(g\in G\) carrying a
fixed reference hidden-state representation \(\cs{g}\in\mathbb F^d\). For a sequence \(s=x_1\cdots x_T\), let
\(F_s:=F_{x_T}\circ\cdots\circ F_{x_1}\) be the induced hidden-state map, where \(F_x\) is the update \eqref{eq:canonical_state} under input \(x\). We
focus on \emph{state-preserving} sequences, whose symbolic action \(\mathcal T_s\) is the
identity: \(\mathcal T_s(g)=g\) for all \(g\in G\). Any exact realization must therefore
return every \(\cs{g}\) to itself, \(F_s(\cs{g})=\cs{g}\) for all \(g\in G\).

The perturbations that matter most for symbolic tracking are those that move a
hidden state toward competing representations \(\cs{g'}\). These directions span the
\emph{symbolic subspace}
\begin{equation}
    \mathcal U:=\operatorname{span}\{\cs{g}-\cs{g'}:g,g'\in G\}.
    \label{eq:symbolic_subspace}
\end{equation}
Thus the directions that separate symbolic states are also the directions along
which errors appear.

\begin{theorem}[Affine neutrality on the symbolic subspace]
\label{thm:affine_neutrality}
Let \(s\) be a state-preserving sequence with non-degenerate representations
(\(\cs{g}\neq \cs{g'}\) for \(g\neq g'\)), and suppose the induced return
map is affine, \(F_s(h)=A_s h+b_s\) for all hidden states \(h\in\mathbb F^d\). If \(F_s(\cs{g})=\cs{g}\) for all
\(g\in G\), then
\[
A_s|_{\mathcal U}=I.
\]
\end{theorem}

\Cref{thm:affine_neutrality} shows that once an affine return map fixes every
symbolic state exactly, it has no freedom left to shrink the directions that
separate those states. For any \(g\in G\) and perturbation
\(\delta\in\mathcal U\),
\[
F_s(\cs{g}+\delta)-F_s(\cs{g})=\delta.
\]
Thus symbolic realization and symbolic correction are incompatible on
\(\mathcal U\): exact affine models may preserve every \(\cs{g}\), but they
cannot create a restoring attractor along the directions that matter for
symbolic discrimination. Proofs are in \Cref{ax:proof}, which also extends the theorem to non-minimal realizations, where several hidden states represent one symbolic state (\Cref{proof:nonminimal}).

\paragraph{State-dependent error correction.}
On the other hand, a state-dependent return map can fix each representation $\cs{g}$
without being neutral around it. Writing a perturbed state as
\(\cs{g}+p\) with \(p\in\mathcal U\), the relevant local map is
\(p\mapsto F_s(\cs{g}+p)-\cs{g}\). If its Jacobian at \(p=0\) has norm
strictly below one uniformly over \(g\), then nearby symbolic-subspace
errors contract and every $\cs{g}$ is locally attracting. Thus state
dependence does not guarantee correction, but it permits the
state-conditioned perturbation contraction that affine return maps
cannot realize; \Cref{ax:nonlinearity_jacobians} works out which
choices of nonlinearity \(\phi\) deliver this Jacobian-contraction
condition operationally.

\subsection{Accumulated error controls finite-horizon tracking}
\label{subsec:theory_error_budget}

\Cref{thm:affine_neutrality} does not imply immediate failure: affine return
dynamics cannot generically remove errors along state-separating directions.
The finite-horizon question is how long the learned symbolic states remain
distinguishable under repeated reuse.

Let \(c_g(t):=\mathbb{E}[h_t\mid g_t=g]\) denote the empirical centroid of the trained model's hidden
states with symbolic state \(g\) at step \(t\); unlike the fixed reference \(c^\star_g\), it may drift over the rollout. Let \(W_{\mathrm{out}}\in
\mathbb{F}^{|G|\times d}\) be the linear readout the classifier reads
from. Define readout-space quantities
\[
R(t):=\mathbb{E}\bigl[\|W_{\mathrm{out}}(h_t-c_{g_t}(t))\|_2\bigr],
\quad
M(t):=\min_{g\neq g'}\|W_{\mathrm{out}}(c_g(t)-c_{g'}(t))\|_2,
\quad
q(t):=R(t)/M(t).
\]
\(R(t)\) is the within-class spread the decoder sees, \(M(t)\) is the
between-class separation, and \(q(t)\) is the distinguishability ratio. We call the nearest-centroid bound \(\tau = \tfrac12\) the
\emph{readability threshold}: symbolic states remain readable while \(q(t)<\tau\).
Further, let \(P_{\mathcal U}\) denote the
orthogonal projection onto the symbolic subspace \(\mathcal U\).
For an exact state-preserving return map \(F_s(h)=A_s h+b_s\), \Cref{thm:affine_neutrality}
implies \(A_s|_{\mathcal U}=I\).

\begin{corollary}[Finite-horizon error accumulation]
\label{cor:error_budget_modes}
Let the trained return-cycle tracker be
\(\tilde F_s=F_s+\varepsilon\), where \(F_s\) is the exact state-preserving affine return map considered
in \Cref{thm:affine_neutrality}. Along a return cycle,
let \(h_t\) be the trained model's hidden state at step \(t\) and \(c^\star_{g_t}\) the fixed reference representation of its ground-truth symbolic state \(g_t\);
define \(e_{\mathcal U}(t):=P_{\mathcal U}(h_t-\cs{g_t})\) and
\(\eta_t:=P_{\mathcal U}\varepsilon(h_t)\). Then
\[
e_{\mathcal U}(t)
=
e_{\mathcal U}(0)+\sum_{j=0}^{t-1}\eta_j .
\]
Thus any coherent residual component accumulates linearly.
If, over the relevant horizon, the projected residuals have a nonzero
average drift
\(t^{-1}\sum_{j<t}\eta_j\approx\bar\eta\neq0\) and \(M(t)\approx M>0\), then \(q(t)\)
crosses a fixed threshold \(\tau\) on the scale
\[
T_{\mathrm{cross}}
\approx
\frac{\tau M}{\|W_{\mathrm{out}}\bar\eta\|}.
\]
\end{corollary}

Proof in \Cref{app:error_budget}. Empirically, the affine
models we test (\Cref{subsec:exp_codebook_collapse}) trace out two
trajectories of \(q(t)\) consistent with this picture:
\emph{saturation}, where \(q(t)\) sits above \(\tau\) from the first
few steps because \(R(0)/M(0)\) is already large, and \emph{climb},
where \(q(t)\) starts below \(\tau\) and grows linearly until the
crossing, exactly the regime in which the
\(T_{\mathrm{cross}}\) estimate
above applies.

\section{Experiments}
\label{sec:exp}

\begin{table}[t]
    \centering
    \small
    \begin{tabular}{l l  cc cc cc}
        \toprule
        & & \multicolumn{2}{c}{$C_2$} & \multicolumn{2}{c}{$C_6$} & \multicolumn{2}{c}{$S_3$} \\
        \cmidrule(r){3-4} \cmidrule(r){5-6} \cmidrule{7-8}
        Model & Dynamics & L1 & L2 & L1 & L2 & L1 & L2 \\
        \midrule
        Mamba           & Affine          & \bad &   60 & \bad &   60 & \bad & \bad \\
        Mamba-3         & Affine          &  200 &  300 &  100 &  100 & \bad &   60 \\
        AUSSM           & Affine          & 1000 & \bad &  200 &  100 & \bad & \bad \\
        Simple AUSSM    & Affine          &  300 &  400 &  100 &  100 &   60 &  100 \\
        Negative Mamba  & Affine          & 1000 & 1000 &  100 &  200 &  100 &  200 \\
        Linear RNN      & Affine          & \bad &  100 & \bad &   60 & \bad & \bad \\
        Token-gated RNN & Affine          & 1000 &  700 &  300 &  400 &  500 & 1000 \\
        \midrule
        Gated DeltaNet     & Affine & 1000 & 1000 & \bad & 100 & \bad & 200 \\
        KDA                & Affine & 1000 & 1000 & \bad & 100 & \bad & 200 \\
        Gated DeltaNet-2   & Affine & 1000 & 1000 & \bad & 100 & \bad & 200 \\
        Gated DeltaProduct & Affine & 1000 & 1000 & 400  & 200 & 400  & 200 \\
        \midrule
        tanh RNN        & State-dependent & 1000 & 1000 & 1000 & 1000 & 1000 & 1000 \\
        State-gated RNN & State-dependent & 1000 & 1000 & 1000 & 1000 & 1000 & 1000 \\
        \bottomrule
    \end{tabular}
    \vspace{0.5em}
    \caption{\textbf{Model performance on state-tracking tasks.} Values indicate the max-passing length~$\mathrm{mp}$, the largest evaluation length at which test accuracy remains $\geq 90\%$. The maximum training length was $60$; cells displaying $60$ denote models that survive only at the curriculum length. \bad~denotes $\mathrm{mp}=0$ (the curriculum did not converge). L1 and L2 denote one-layer and two-layer recurrent stacks.}
    \label{tab:main}
\end{table}

\paragraph{Models.}
We evaluate recurrent architectures spanning a spectrum from SSMs to gated RNNs. The \emph{affine} models are: \textit{Mamba}~\citep{gu2024mamba}, a selective SSM; \textit{Mamba-3}~\citep{lahoti2026mamba3}, a more expressive SSM variant; \textit{AUSSM}~\citep{karuvally2025aussm}, an adaptive unitary SSM; \textit{Simple AUSSM}~\citep{shakerinava2026expressive}, an ablated AUSSM variant; \textit{Negative Mamba}~\citep{grazzi2024unlocking}, a Mamba variant with signed transition factors; \textit{Linear RNN}, a dense real-valued linear recurrence; and \textit{Token-gated RNN}, a gated recurrence whose gate depends only on the input \(x_t\), so the update remains affine in \(h_{t-1}\). We also evaluate four delta-rule models whose matrix-valued state update is affine in the state for a given input: \textit{Gated DeltaNet}~\citep{yang2025gated}, \textit{KDA}~\citep{kimiteam2025kimilinear}, \textit{Gated DeltaNet-2}~\citep{hatamizadeh2026gateddeltanet2}, and \textit{Gated DeltaProduct}~\citep{siems2025deltaproduct}. The \emph{state-dependent} models are: \textit{tanh RNN}, a standard Elman RNN~\citep{elman1990finding} with \(\tanh\) activation; and \textit{State-gated RNN}, a simplified gated recurrence whose gate depends on both \(h_{t-1}\) and \(x_t\). \Cref{ax:model} gives the full operator definitions.

\paragraph{Tasks.}
We evaluate on three group state-tracking tasks of increasing difficulty: parity $C_2$, the cyclic group $C_6$, and the symmetric group $S_3$. Each task requires tracking the running group product $g_t = g_{t-1} \cdot x_t$ from input elements sampled uniformly from the group. 
Detailed explanation and examples are in \Cref{ax:task}.

\paragraph{Experimental \texorpdfstring{details}{details}.}
All models are trained with curriculum learning up to sequence length \(60\). We evaluate extrapolation at lengths \(\{100,200,\ldots,1000\}\) and report the maximum length at which test accuracy remains at least \(90\%\). For each model and task, we run a grid search over state dimension, learning rate, learning-rate schedule, and three random seeds, reporting the best-performing configuration (full grid in \Cref{tab:exp_hparams}). We evaluate both single-layer and two-layer recurrent stacks, denoted L1 and L2. Additional training and evaluation details are provided in \Cref{ax:exp}.

\subsection{\texorpdfstring{State-tracking performance}{State-tracking performance}}
\label{subsec:exp_downstream}

\Cref{tab:main} reveals a clear dichotomy in state-tracking robustness: state-dependent models (tanh RNN and State-gated RNN) reliably track symbolic states up to the maximum tested length of \(1000\) tokens across all three tasks (\(C_2\), \(C_6\), and \(S_3\)), whereas affine models are generally unstable, with a few exceptions: Negative Mamba and the delta-rule models on \(C_2\), and Token-gated RNN on \(C_2\) and \(S_3\).
The results match \Cref{thm:affine_neutrality}: recurrent operators
without state-dependent transitions lack robust error correction.

At the same time, the results show that some affine operators can remain on track well beyond the training length of \(60\) tokens. For example, Negative Mamba reaches \(1000\) on \(C_2\), and Token-gated RNN reaches \(1000\) on \(C_2\) (L1) and \(S_3\) (L2), with shorter horizons of \(500\) on \(S_3\) (L1) and \(400\) on \(C_6\) (L2). These cases show that affine dynamics can approximate correction over finite horizons, and in two settings extend tracking out to the maximum tested length.

\subsection{Error control behavior}
\label{subsec:exp_perturbation}

\begin{figure}[t]
    \centering
    \includegraphics[width=0.98\linewidth]{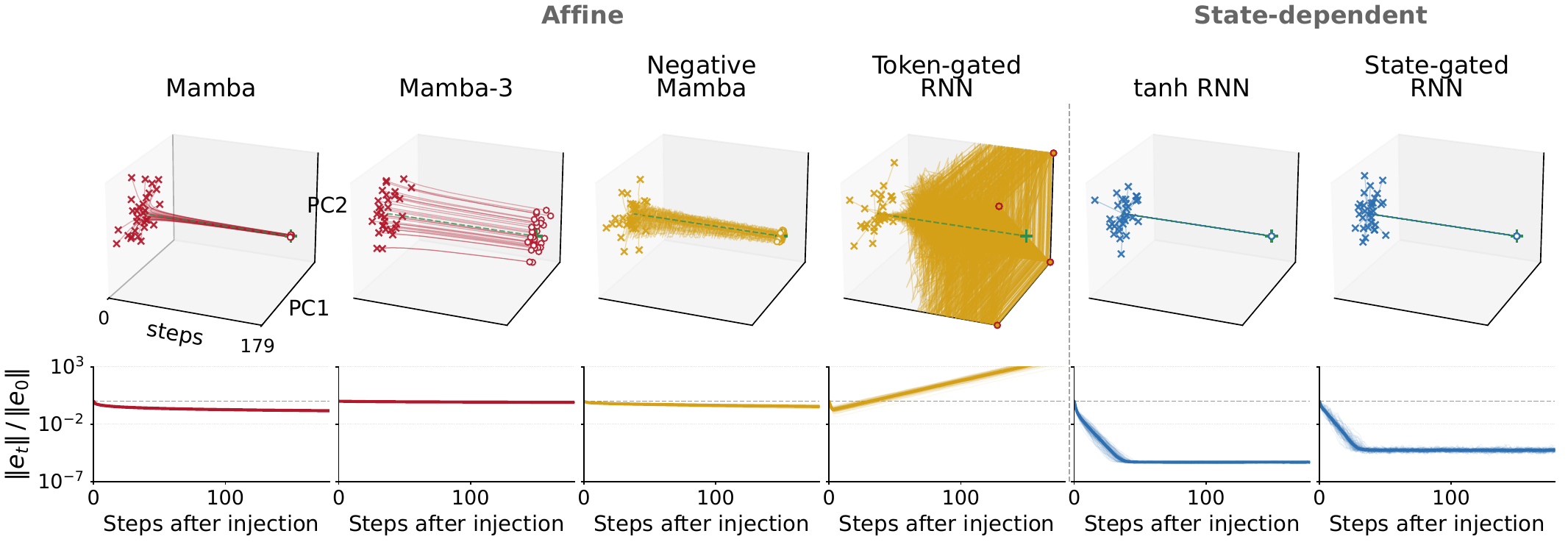}
    \caption{\textbf{Perturbation recovery after noise injection on $S_3$.} Top: error trajectories in a fixed 2D PCA basis of the injected errors; $\times$ marks the injected error, $\circ$ the endpoint (filled when clipped off scale), and the green dashed line and $+$ zero error. Bottom: normalized error magnitude $\|e_t\|/\|e_{t_0}\|$ over time, median and interquartile range over trials; thin lines show individual trials, and the dashed line a ratio of $1$. Affine models show either global decay (Mamba, Mamba-3, and Negative Mamba) or expansion (Token-gated RNN), while state-dependent models (tanh RNN and State-gated RNN) show strong, selective error contraction.}
    \label{fig:perturbation}
\end{figure}

Next, we test each model's error control dynamics, as predicted by~\Cref{thm:affine_neutrality}. We inject a hidden-state perturbation and measure whether the error is propagated or reduced.

\paragraph{Metric.}
Error correction is operationally the decay of an injected perturbation
under propagation.
We inject Gaussian noise at step \(t_0=20\) and compare the perturbed rollout with a clean rollout under the same input sequence $i$. Given the stepwise hidden states $h^{\mathrm{pert}}_{i,t}$ and $h^{\mathrm{clean}}_{i,t}$, we measure
\[
e_{i,t}
=
h^{\mathrm{pert}}_{i,t}
-
h^{\mathrm{clean}}_{i,t},
\qquad
\mathrm{ratio}_{i,t}
=
\frac{\|e_{i,t}\|_2}{\|e_{i,t_0}\|_2}.
\]
We track the full hidden-state difference rather than its projection onto
$\mathcal U$, since the goal here is to characterize each model's overall
response to perturbation; symbolic-subspace dynamics are addressed
separately in \Cref{subsec:exp_subspace_decomp}.

\paragraph{Results.}
\Cref{fig:perturbation} shows the accumulated response to injected
perturbations. A clear dichotomy emerges. State-dependent models (tanh
RNN, State-gated RNN) collapse $\|e_t\|/\|e_{t_0}\|$ by several orders of
magnitude within tens of steps and hold near the floor. The affine SSMs
(Mamba, Mamba-3, Negative Mamba) instead contract errors through their
global diagonal decay $\alpha_t$ with $|\alpha_t|<1$, at
per-step rates $\rho_{\mathrm{step}}<1$. This indicates global dissipation
rather than conditional error correction for the affine models.

Token-gated RNN amplifies perturbations (\(\rho_{\mathrm{step}}>1\)), with
\(\|e_t\|/\|e_{t_0}\|\) reaching orders of magnitude above one. This follows from
\(h_t=g(x_t)\odot W h_{t-1}+W_x x_t+b\), where
\(g(x_t)=\sigma(W_gx_t+b_g)\): because the gate depends only on \(x_t\),
clean and perturbed rollouts share the same gates, so errors follow
\(e_t=g(x_t)\odot W e_{t-1}\). Mamba variants have the same cancellation but
dissipate through \(|\alpha_t|<1\); Token-gated RNN instead relies on a dense
\(W\) with spectral radius \(\ge 1\) to keep group states separable, which also
amplifies errors.

\subsection{State separation over rollouts}
\label{subsec:exp_codebook_collapse}

\begin{figure}[t]
    \centering
    \includegraphics[width=0.98\linewidth]{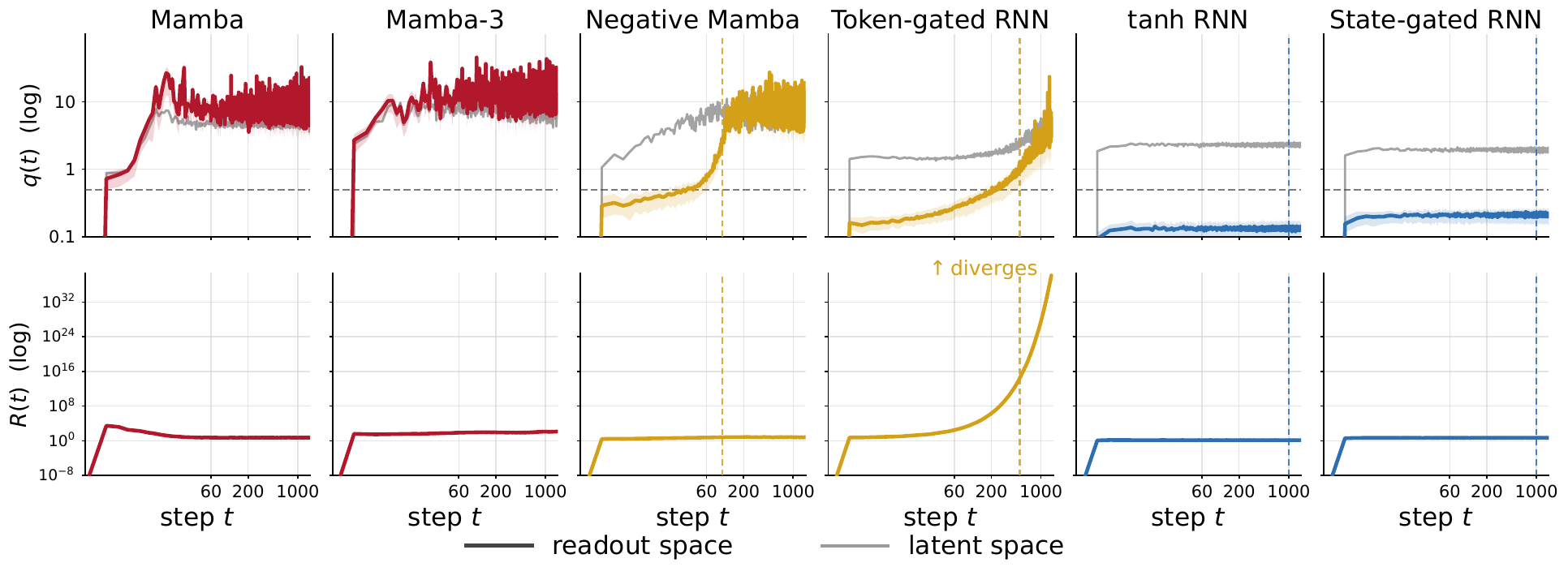}
\caption{\textbf{Distinguishability ratio $q(t)$ over rollouts on $S_3$.}
Top: $q(t)=R(t)/M(t)$ from \Cref{subsec:theory_error_budget}, with the
dashed line marking the nearest-centroid bound $q(t)=1/2$; gray curves show the latent-space $q(t)$. Bottom:
$R(t)$ on log-log axes.
Vertical dashes mark each model's one-layer $S_3$ max-passing length from \Cref{tab:main}.
Means over $N=200$ rollouts; IQR shown as same-color band. Affine models show either immediate saturation (Mamba,
Mamba-3) or gradual climb (Negative Mamba and Token-gated RNN), while
state-dependent models (tanh RNN and State-gated RNN) remain low.}
    \label{fig:codebook_collapse}
\end{figure}

Here, we directly put the framework of \Cref{subsec:theory_error_budget} to test.
\Cref{subsec:theory_error_budget} describes two failure modes for an approximate
affine tracker: \emph{saturation}, where $q(t)$ sits above the readability
threshold $\tau$ from the start, or \emph{climb}, where $q(t)$ starts below
$\tau$ and crosses it at $T_{\mathrm{cross}}$. We measure $q(t)$ across
rollouts and inspect how each architecture's trajectory unfolds against
these predictions.

\paragraph{Metric.}
At each step $t$ we form time-current centroids
$c_g(t):=\mathbb{E}_i[h_{i,t}\mid g_{i,t}=g]$, the per-step mean of
the hidden state over $N=200$ rollouts whose oracle symbol at $t$ is
$g$. We then measure the distinguishability ratio $q(t)=R(t)/M(t)$
from \Cref{subsec:theory_error_budget}: $R(t)$ is the empirical mean
over rollouts of $\|W_{\mathrm{out}}(h_{i,t}-c_{g_{i,t}}(t))\|_2$, and $M(t)$ is the
smallest pairwise centroid distance in readout space, $\min_{g\neq g'}\|W_{\mathrm{out}}(c_g(t)-c_{g'}(t))\|_2$.
Thus $q(t)$ measures within-class spread in units of the smallest
inter-class margin. For a nearest-centroid decoder, $q(t)<0.5$ means that a state at
the average distance $R(t)$ is still closer to its own centroid than to
any competitor.

\paragraph{Results.}
\Cref{fig:codebook_collapse} supports the two structural predictions
of \Cref{thm:affine_neutrality} and \Cref{cor:error_budget_modes}. In the top
row, state-dependent transitions (tanh RNN and State-gated RNN)
maintain \(q(t) < 0.5\) throughout rollout, whereas affine transitions
eventually cross the nearest-centroid bound, consistent with the
no-correction obstruction of \Cref{thm:affine_neutrality}.

Within the affine class, the trajectories instantiate the two
finite-horizon alternatives of \Cref{subsec:theory_error_budget}. Mamba
and Mamba-3 are already above the nearest-centroid bound at the start of
extrapolation, corresponding to saturation: the state clouds are
immediately too wide relative to their separation. Negative Mamba and
Token-gated RNN follow the climb regime: they start below the
bound, remain readable for a finite horizon, and cross only after
repeated rollout accumulates readout-space spread. In both climb
cases, the crossing precedes the corresponding \(\mathrm{mp}\) in
\Cref{tab:main}, consistent with a readability-based failure
criterion.

The bottom row shows that the climb regime can arise through different
dynamics of spread and separation. Token-gated RNN grows \(R\) together
with \(M\), keeping the latent-space ratio comparatively stable.
Negative Mamba instead directly bounds \(R\) through its diagonal
transition parameterization, yielding a slower climb despite the same
affine no-correction constraint.

\subsection{Error decomposition along the symbolic subspace}
\label{subsec:exp_subspace_decomp}

\begin{figure}[t]
    \centering
    \includegraphics[width=0.98\linewidth]{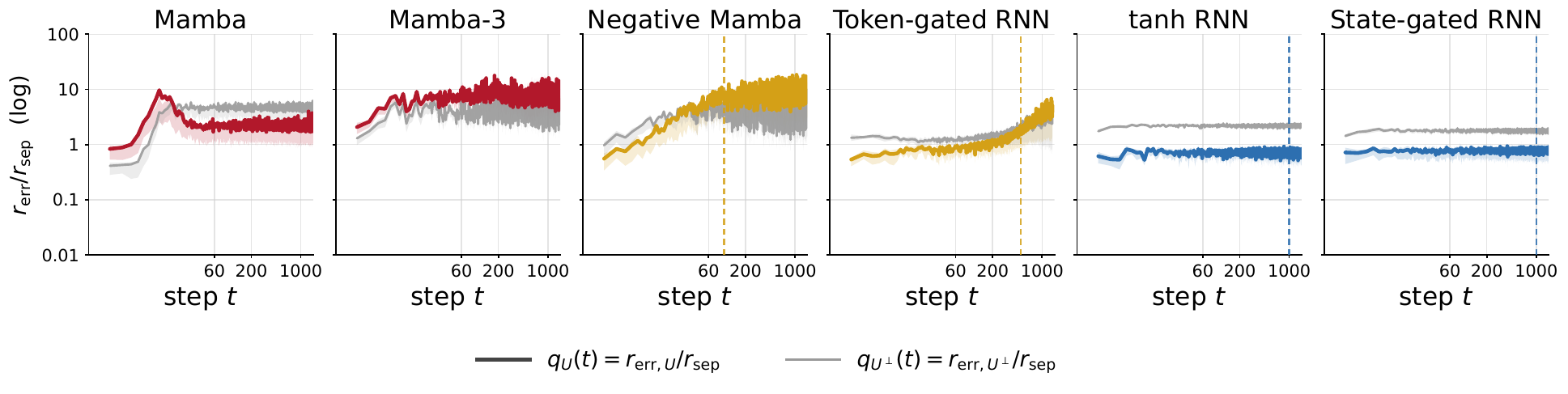}
    \caption{\textbf{Subspace decomposition of within-class spread.}
For each architecture, we decompose within-class spread into the
symbolic-subspace component \(q_{\mathcal U}(t) =
r_{\mathrm{err},\mathcal U}(t)/r_{\mathrm{sep}}(t)\) (color) and the
orthogonal component \(q_{\mathcal U^\perp}(t) =
r_{\mathrm{err},\mathcal U^\perp}(t)/r_{\mathrm{sep}}(t)\) (gray),
both on a log scale. Vertical dashes mark each model's one-layer \(S_3\) max-passing length
from \Cref{tab:main}.}
    \label{fig:subspace_decomp}
\end{figure}

We next ask whether the deviation lies in the
symbolic subspace \(\mathcal U\) from \Cref{eq:symbolic_subspace},
where symbolic errors appear and affine return dynamics cannot
generically contract perturbations (\Cref{thm:affine_neutrality}). We
therefore decompose the within-class deviation into \(\mathcal U\) and
\(\mathcal U^\perp\) components.

\paragraph{Metric.}
At step \(t\), let \(\delta_{i,t}:=h_{i,t}-c_{g_{i,t}}(t)\) be the
per-rollout deviation from the time-current centroid, and let
\(P_{\mathcal U}(t)\) project onto the span of centroid differences
\(\{c_g(t)-c_{g'}(t)\}\).
Define the root-mean-square spreads
\[
r_{\mathrm{err},\mathcal U}(t)
:=\sqrt{\mathbb{E}_i\|P_{\mathcal U}(t)\,\delta_{i,t}\|_2^2},
\qquad
r_{\mathrm{err},\mathcal U^\perp}(t)
:=\sqrt{\mathbb{E}_i\|\delta_{i,t}\|_2^2-r_{\mathrm{err},\mathcal U}(t)^2},
\]
and the inter-centroid scale
\(r_{\mathrm{sep}}(t):=\min_{g\neq g'}\|c_g(t)-c_{g'}(t)\|_2\). We
report \(q_{\mathcal U}(t)=r_{\mathrm{err},\mathcal U}(t)/r_{\mathrm{sep}}(t)\)
and \(q_{\mathcal U^\perp}(t)=r_{\mathrm{err},\mathcal U^\perp}(t)/r_{\mathrm{sep}}(t)\),
which split the within-class spread into state-separating and
orthogonal components. RMS aggregation preserves the per-rollout
Pythagorean identity at the population level.

\paragraph{Results.}
\Cref{fig:subspace_decomp} shows how the spread is distributed across
\(\mathcal U\) and \(\mathcal U^\perp\). For Negative Mamba and
Token-gated RNN, \(q_{\mathcal U^\perp}\) is larger than
\(q_{\mathcal U}\) early in rollout, indicating that most spread
initially lies outside the state-separating directions. Near each
model's max-passing length the ordering reverses: \(q_{\mathcal U}\)
catches up to and exceeds \(q_{\mathcal U^\perp}\). Thus,
finite-horizon failure is associated not merely with growth of spread,
but with its shift into \(\mathcal U\), the subspace where affine
return dynamics cannot generically contract perturbations.

State-dependent models (tanh RNN and State-gated RNN) show the
complementary pattern: \(q_{\mathcal U}\) remains suppressed while the
larger component lies in \(\mathcal U^\perp\). State-dependent
transitions selectively prevent spread along the state-separating
directions, supplying the conditional correction unavailable to affine
return maps under \Cref{thm:affine_neutrality}. Mamba and Mamba-3 are
saturated from early rollout, so there is no meaningful
subspace-dominance transition to analyze.

\subsection{Further analysis}
\label{subsec:exp_analysis}

\paragraph{Additional models and tasks.}
We report additional results in \Cref{subsec:ax_more_models,subsec:ax_more_tasks}. \Cref{subsec:ax_delta_rule} repeats the analyses of Figures~\labelcref{fig:perturbation}, \labelcref{fig:codebook_collapse}, and~\labelcref{fig:subspace_decomp} for Gated DeltaProduct, the only delta-rule model whose curriculum converges on \(S_3\) with one layer.

\begin{figure}[t]
  \centering
  
  \begin{minipage}{0.46\textwidth}
    \centering
    \setlength{\tabcolsep}{4pt}
    \begin{tabular}{c l c}
        \toprule
        Type & Operator \(\phi\) & \(S_3\) \\
        \midrule
        & Affine & \bad \\
        \midrule
        \multirow{2}{*}{\rotatebox[origin=c]{90}{norm}}
            & LayerNorm & \bad \\
            & sphere projection & \bad \\
        \midrule
        \multirow{5}{*}{\rotatebox[origin=c]{90}{nonlinear}}
            & \(\tanh\) & 1000 \\
            & ReLU & 1000 \\
            & max & 1000 \\
            & min & 1000 \\
            & GroupSort $k=2$ & 1000 \\
        \bottomrule
    \end{tabular}
    \vspace{4mm}
    \captionof{table}{\textbf{Many nonlinear operators support robust tracking.}
On \(S_3\) in the single-layer setting, diverse nonlinear activations reach the maximum tested
length, whereas affine and normalization-only variants fail.
Refer to \Cref{ax:nonlinearity_jacobians} for interpretation.}
    \label{tab:nonlinearities}
  \end{minipage}\hfill %
  \begin{minipage}{0.50\textwidth}
    \centering
    \includegraphics[width=\linewidth]{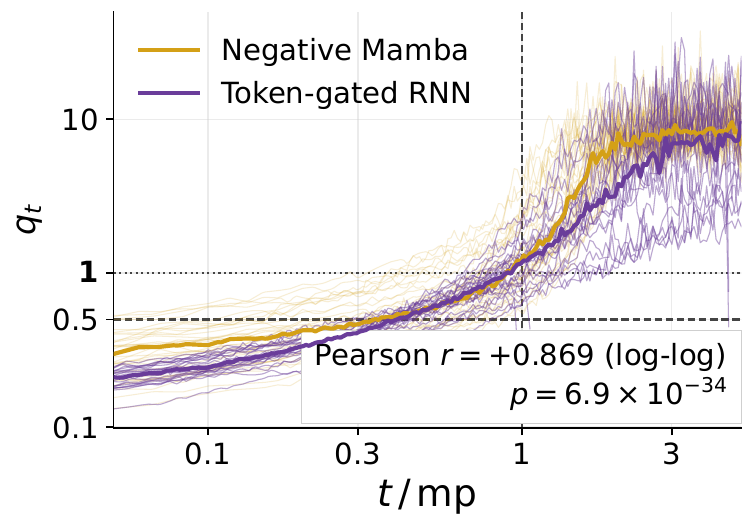}
    \vspace{-6mm}
    \caption{\textbf{Readability collapse coincides with downstream failure.}
On $S_3$, the distinguishability ratio $q(t)$ is plotted against
failure-normalized time $t/\mathrm{mp}$. Thin lines are individual runs, thick lines the median of each architecture, and both axes are logarithmic. The dotted reference at
$q(t)=1$ marks where within-class spread equals the smallest inter-class
margin. The dashed lines mark the threshold $q(t)=1/2$ and $t=\mathrm{mp}$. Pearson $r=0.87$ on log $T_{\mathrm{cross}}$ vs log $\mathrm{mp}$.
}
    \label{fig:corr}
  \end{minipage}
\end{figure}

\paragraph{Correlation between \(T_{\mathrm{cross}}\) and downstream performance.}
\Cref{cor:error_budget_modes} predicts that the first nearest-centroid
crossing, \(T_{\mathrm{cross}}=\min\{t:q(t)\geq 0.5\}\) (\Cref{ax:tcross}), should track how
long an affine tracker remains usable. \Cref{fig:corr} supports this:
across \(S_3\) models with \(\mathrm{mp}\geq 60\) that cross,
\(T_{\mathrm{cross}}\) strongly correlates with downstream max-passing
length on a log-log scale (\(r=+0.87\), \(p<10^{-30}\)).

The trained readout fails later than the nearest-centroid bound
suggests: although the nearest-centroid bound is \(q(t)<0.5\), failure empirically aligns closer to \(q(t)=1\), where
within-class spread matches between-class separation. At \(t=\mathrm{mp}\),
the median \(q(t)\) is \(0.97\) (\(95\%\) bootstrap CI
\([0.84,\,1.08]\)). \(T_{\mathrm{cross}}\) remains predictive because both
thresholds are driven by the accumulated within-class spread predicted by
\Cref{cor:error_budget_modes}.

\paragraph{Nonlinear activation type.}
\Cref{thm:affine_neutrality} identifies state-dependent transitions as the key ingredient for error correction, not a specific nonlinear implementation. To test this, we fix the Elman recurrence \(h_t=\phi(W_hh_{t-1}, W_xx_t+b)\) and vary only the nonlinear activation \(\phi\).

\Cref{tab:nonlinearities} shows that several distinct state-dependent operators succeed, including standard pointwise activations, pointwise \(\max/\min\), and GroupSort~\citep{anil2019sorting}. In contrast, whole-vector normalization operators fail despite being nonlinear. Thus the relevant distinction is not the activation family itself, but whether the induced Jacobian can modulate symbolic directions in a state-dependent way. We defer the operator-level Jacobian analysis to \Cref{ax:nonlinearity_jacobians}.

\paragraph{\(C_2\) is a weak test of correction.}
Several affine models reach the maximum tested length on \(C_2\) because
parity can remain readable under neutral oscillation, without genuine error
correction. With \(C_2=\{e,a\}\), the affine involution \(F_a(h)=-h+(\cs{e}+\cs{a})\) swaps any two
distinct fixed references and satisfies \(F_a^2=\mathrm{id}\). For a
symbolic-subspace perturbation,
\[
F_a(\cs{g}+\delta)=\cs{g\cdot a}-\delta,\qquad
F_a^2(\cs{g}+\delta)=\cs{g}+\delta,
\]%
so the error flips sign but is not removed. Binary decoding can still
remain correct while this oscillation stays within the readout margin.
Thus \(C_2\) is an order-two edge case: it tests whether an uncorrected error stays within the readout margin,
not active correction of symbolic-subspace drift. See
\Cref{app:parity_edge_case}.
\section{Conclusion}
\label{sec:conclusion}

Recurrent state tracking is usually framed as expressivity: which symbolic
transitions an architecture can represent. We show that this is incomplete:
robust tracking also requires controlling errors accumulated under repeated
reuse. Affine models cannot correct errors on the state-separating subspace, because preserving every symbolic state forces the identity action on the directions that separate the states
(\Cref{thm:affine_neutrality}). Thus approximate affine trackers fail when
accumulated error overtakes the state margin, not when expressivity runs out.
Empirically, affine models saturate or climb past the readability threshold,
whereas state-dependent models retain conditional contraction and track far
beyond the training length. Overall, state tracking is limited
not only by what a model can represent, but by whether it can correct the
errors accumulated over time.

Therefore, future work on architectures for state tracking could explore
state-dependent correction, for example through nonlinear RNNs with parallel
training~\citep{gonzalez2024scalable,danieli2026pararnn} or hybrids that
interleave affine layers with attention or nonlinear
layers~\citep{kimiteam2025kimilinear}; a state-dependent update gives up the
exact parallel scan that makes affine models efficient.
Limitations are
discussed in \Cref{ax:lim}.

\begin{ack}
This work was supported by Institute of Information \& Communications Technology Planning \& Evaluation (IITP) grants funded by the Korea government (MSIT) (No. RS-2020-II201361, Artificial Intelligence Graduate School Program (Yonsei University); No. RS-2022-II220113, Developing a Sustainable Collaborative Multi-modal Lifelong Learning Framework) and the National Research Foundation of Korea (NRF) grant funded by the Korea government (MSIT) (RS-2025-00554790).
\end{ack}

{
\small
\bibliographystyle{plainnat}
\bibliography{references}
}

\clearpage
\appendix
\crefalias{section}{appendix}\crefalias{subsection}{appendix}

\section*{Appendix Overview}
\label{ax:overview}

\begin{itemize}
    \item \textbf{\Cref{ax:lim}:} Limitations
    \item \textbf{\Cref{ax:rel}:} Related work
    \item \textbf{\Cref{ax:exp}:} Experimental details
    \item \textbf{\Cref{ax:proof}:} Proofs
    \item \textbf{\Cref{ax:further_discussions}:} Further discussion
    \item \textbf{\Cref{ax:more_results}:} Additional empirical results
    \item \textbf{\Cref{ax:task}:} Examples of state-tracking tasks
    \item \textbf{\Cref{ax:model}:} Model details
\end{itemize}

\section{Limitations}
\label{ax:lim}

This work isolates how recurrent dynamics manage error during symbolic state
tracking. Our results do not imply that affine recurrences fail in-domain, nor
that they are unsuitable for sequence modeling in general. They show that, when
a model must reuse symbolic states beyond the training horizon, affine return
maps cannot provide the state-dependent error dynamics needed to keep
accumulated drift from erasing state separation.

Our experiments are restricted to finite-group state tracking, the canonical
testbed used in prior theoretical work on recurrent state
tracking~\citep{merrill2024illusion,sarrof2024expressive,shakerinava2026expressive}.
The main experiments use $C_2$, $C_6$, and $S_3$, and the appendix extends the
suite to $C_2\!\times\!C_4$ and $A_4$; the affine/state-dependent dichotomy
persists across all five groups tested (\Cref{tab:main,tab:more_tasks}).
This restriction is deliberate. The tasks are difficult enough to expose
failure, yet simple enough that architectures do not collapse into uniform
failure. This separation makes differences in correction dynamics directly
observable, whereas richer benchmarks would add confounds and may obscure the
mechanism behind failure.
The capability these tasks isolate, maintaining and updating an exact
latent state over long horizons, is one that many real-world problems also
require, such as tracking variables and scopes in long-form code generation or
character identities in a video narrative. Group tracking is thus a necessary test for such problems, not a sufficient
one. Language-like tasks such as bracket matching, code execution, or
board-game move sequences are a natural next step; they need an exactly
identifiable intermediate state and little confounding from parsing or task
knowledge, and we leave them to future work.

On the theory side, \Cref{thm:affine_neutrality} assumes an exact affine
state-preserving return map, while the experiments train approximate models.
\Cref{cor:error_budget_modes} bridges the two through the residual
$\varepsilon$, but its drift $\bar\eta$ depends on the action of $A_s$ on
$\mathcal U^\perp$, which the theorem leaves unconstrained, so it cannot be
recovered from a trained model; we therefore test the corollary only through
its observable consequences. The theorem itself covers point-valued
representations and, through \Cref{proof:nonminimal}, non-minimal
realizations whose decoder classes have distinct preserved centroids, but not
decoder classes with coinciding centroids, transient hidden states, or
arbitrary decoder equivalence classes.

Two further limitations concern the experiments. \Cref{tab:main} reports
the best max-passing length across seeds, which does not show how reliably it
is reached; \Cref{subsec:ax_seeds} reports
the mean and standard deviation across seeds. In addition, our analysis is
restricted to
recurrent state-tracking models, following prior work on recurrent state
tracking~\citep{merrill2024illusion,sarrof2024expressive,shakerinava2026expressive}.
Attention-based models are not recurrent state-update models, so the
hidden-state error analysis does not apply to them; \Cref{subsec:ax_transformer}
reports their downstream state-tracking performance under the same task
settings for comparison.

\section{Related \texorpdfstring{work}{work}}
\label{ax:rel}

\paragraph{Recurrent \texorpdfstring{models.}{models.}}
RNNs model sequential data through recurrent hidden-state updates. 
\citet{elman1990finding} introduced an RNN with a nonlinear activation, while LSTM \citep{hochreiter1997lstm} and GRU \citep{cho2014learning} address vanishing gradients with gating mechanisms, yet their nonlinear recurrence limits parallelization.
Recent works therefore revisit linear recurrent architectures for better scalability. LRU \citep{orvieto2023resurrecting} uses diagonal linear dynamics, while DeltaNet \citep{yang2024deltanet} adopts delta-rule-based linear recurrence for efficiency.
Parallel to these, SSMs characterize temporal dynamics by discretizing continuous-time differential equations into linear recurrences.
Similarly, SSM-based models use linear state updates and often employ structured transition matrices for efficiency \citep{gupta2022diagonal, gu2022efficiently}.
Mamba \citep{gu2024mamba} introduces input-dependent selective updates; more recently, Mamba-3 \citep{lahoti2026mamba3} extends this line with complex-valued states and MIMO formulations to improve efficiency and modeling capacity.

\paragraph{Model \texorpdfstring{expressivity and state tracking.}{expressivity and state tracking.}}
Prior works study model expressivity to characterize state-tracking ability in sequence models. 
\citet{merrill2024illusion} show via circuit complexity that linear SSMs with input-independent or diagonal transition matrices, commonly used in recent SSM models, lie in L-uniform $\mathsf{TC}^0$, similar to Transformers. 
Thus, they cannot solve $\mathsf{NC}^1$-hard state-tracking problems such as $S_5$ permutation composition, whereas a single-layer nonlinear RNN can.
\citet{sarrof2024expressive} show that non-negative diagonal SSMs cannot solve parity, revealing a limitation beyond circuit complexity. 
\citet{grazzi2024unlocking} extend this analysis to non-diagonal linear recurrent models, showing that parity requires an eigenvalue outside $[0,\infty)$, such as a negative one, and that counting modulo $m$, for $m$ not a power of two, requires transition products with non-real eigenvalues.

\citet{karuvally2025aussm} introduce AUSSM, an input-dependent complex-valued diagonal SSM with unit-modulus eigenvalues. 
They show that AUSSM can simulate cyclic groups, and that combining it with Mamba enables solvable group recognition, maximizing the expressivity of diagonal SSMs. 
\citet{shakerinava2026expressive} further study the expressivity of input-dependent complex-valued diagonal (DCD) SSMs. 
They show that single-layer DCD SSMs cannot solve state-tracking problems over non-Abelian groups, while multi-layer DCD SSMs can track solvable groups with subnormal series length bounded by the model depth. 
Empirically, however, they also show that DCD SSMs often fail to learn length-generalizing solutions, even for $S_3$ state tracking, despite their theoretical expressivity. 
Motivated by this gap, our work analyzes why such models fail to learn these solutions in practice.

\paragraph{Long-horizon \texorpdfstring{drift in learned latent dynamics.}{drift in learned latent dynamics.}}
\citet{fathi2024course} study long-horizon drift in Koopman autoencoders,
where repeated linear evolution moves the latent state away from the manifold
of valid encoded observations, and correct it by periodically re-encoding the
state. We study drift away from the correct symbolic state: a re-encoded
state can be valid yet represent the wrong symbol, so re-encoding controls
errors in our sense only if it restores the correct symbolic state.

\section{Experimental details}
\label{ax:exp}

\paragraph{Compute.}
All training and analysis runs use one NVIDIA RTX A6000 (48\,GB) or RTX 3090 (24\,GB) GPU per run. The runs of the main grid (\Cref{tab:exp_hparams}) total about $1{,}400$ GPU-hours; a run takes on average from under a minute (tanh RNN) to about two hours (AUSSM). Preliminary experiments not reported here (earlier model variants, other perturbation magnitudes and injection steps, and secondary diagnostics) used a comparable amount of compute; none of them changed the split between affine and state-dependent models.

\paragraph{Statistical reporting.}
Aggregated curves in \Cref{fig:codebook_collapse} show means over
$N=200$ rollouts, and those in \Cref{fig:subspace_decomp} show
root-mean-square spreads over the same rollouts, both with
per-step IQR bands. The
perturbation-recovery panels in \Cref{fig:perturbation} use medians
over $n=200$ injection trials. The $T_{\mathrm{cross}}$ correlation
in \Cref{subsec:exp_analysis} reports the Pearson coefficient with
its $p$-value, together with a $95\%$ bootstrap confidence interval
on the median $q(t)$ at the max-passing length. The max-passing lengths
in \Cref{tab:main} follow the convention of
\citet{shakerinava2026expressive} in reporting the best across three seeds
rather than averaging; \Cref{subsec:ax_seeds} reports the mean and standard
deviation across seeds.

\paragraph{Asset licenses.}
All architectures evaluated in this work are credited via the
citations in \Cref{subsec:ax_more_models,ax:model}. We use PyTorch, NumPy, Matplotlib and the official Mamba implementation
under their published open-source
licenses. The group state-tracking sequences and labels are generated
synthetically by our code.

\paragraph{LLM usage.}
Large language models were used to assist with adapting reference
evaluation code to our experimental setup, generating visualization
scripts, and checking factual consistency of the exposition and
analytic derivations.

\subsection{Training and grid search}
\label{ax:exp_training}

All recurrent models are trained from scratch on group state tracking. Each
model uses the canonical layer of \Cref{def:recursive_layer} inside a pre-norm
residual block, an embedding width $d_{\mathrm{model}}=698$ matched to the
parameter budget of~\citet{shakerinava2026expressive}, and a linear readout
$W_{\mathrm{out}}\in\mathbb{R}^{|G|\times d_{\mathrm{model}}}$. Inputs are
i.i.d. token sequences $x_1,\ldots,x_T$ drawn uniformly from $G$. The model
predicts the running product at every step and is trained with cross-entropy.

Optimization uses AdamW with weight decay $0.01$ and batch size $256$. At each
curriculum stage, we regenerate $10\,000$ training sequences and $2\,000$ test
sequences. The curriculum starts at $T=2$ and doubles whenever test accuracy
is at least $0.95$ for five consecutive epochs, up to $L_{\max}=60$.

After training, we freeze the model and evaluate it on $2\,000$ fresh sequences
for each length in $\{100,200,\dots,1000\}$. The max-passing length
is the largest evaluated length with test accuracy at least
$0.90$. If training reaches $L_{\max}=60$ but no generalization length passes,
we set it to $60$; if the curriculum does not converge, we set
it to $0$ and print it as \bad.

We grid-search the state dimension $d_{\mathrm{state}}$, the learning
rate, the scheduler and the seed over the values in \Cref{tab:exp_hparams}, following
\citet{shakerinava2026expressive}. A \emph{grid-best} checkpoint denotes the
run with the largest max-passing length across all hyperparameter settings
and seeds, with ties broken by the final test accuracy.

\begin{table}[h]
    \centering
    \small
    \begin{tabular}{@{}p{0.42\linewidth}p{0.54\linewidth}@{}}
        \toprule
        \multicolumn{2}{@{}l}{\emph{Architecture}} \\
        \midrule
        $d_{\mathrm{model}}$            & $698$ \\
        depth $L$                       & $\{1, 2\}$ \\
        readout                         & $W_{\mathrm{out}}\in\mathbb{R}^{|G|\times d_{\mathrm{model}}}$ \\
        \midrule
        \multicolumn{2}{@{}l}{\emph{Optimization}} \\
        \midrule
        optimizer                       & AdamW, weight decay $0.01$ \\
        batch size                      & $256$ \\
        total max epochs \newline (across all curriculum stages) & $500$ \\
        gradient clipping         & norm $1$ \\
        cosine schedule           & decays to $10^{-5}$ over $500$ epochs \\
        reduce on plateau         & halves the learning rate after $10$ epochs \newline without a gain in test accuracy \\
        \midrule
        \multicolumn{2}{@{}l}{\emph{Curriculum}} \\
        \midrule
        start length                    & $2$ \\
        max training length $L_{\max}$  & $60$ \\
        promotion threshold             & test accuracy $\geq 0.95$ for $5$ epochs \\
        training / test sequences per stage & $10\,000$ / $2\,000$ \newline (regenerated per stage) \\
        \midrule
        \multicolumn{2}{@{}l}{\emph{Length-generalization evaluation}} \\
        \midrule
        evaluated weights         & best reached in training: longest curriculum \newline length, then highest test accuracy \\
        evaluation lengths            & $\{100,200,\ldots,1000\}$ \\
        sequences per length            & $2\,000$ \\
        passing threshold               & test accuracy $\geq 0.90$ \\
        max-passing length        & largest evaluation length still passing; \newline
                                          $L_{\max}{=}60$ if curriculum converged but no evaluation length passed; \newline
                                          $0$ if curriculum failed \\
        \midrule
        \multicolumn{2}{@{}l}{\emph{Grid search ($81$ runs per model, group and depth)}} \\
        \midrule
        $d_{\mathrm{state}}$            & $\{32, 64, 128\}$ \\
        learning rate                   & $\{10^{-4},\, 5\!\times\!10^{-4},\, 10^{-3}\}$ \\
        scheduler                       & fixed, cosine, reduce on plateau \\
        seeds                           & $3$ \\
        grid-best selection             & largest max-passing length, then final test accuracy \\
        \bottomrule
    \end{tabular}
    \vspace{0.5em}
    \caption{\textbf{Shared training and grid-search configuration.} Used
    for every checkpoint reported in the paper unless explicitly
    overridden in the per-experiment specification.}
    \label{tab:exp_hparams}
\end{table}

\subsection{Per-experiment specification}

\paragraph{State-tracking performance (\Cref{tab:main}).}
The reported number in each cell is the max-passing length of the grid-best
checkpoint for that model, group and depth, evaluated
exactly as described above. No further hyperparameters are introduced.

\paragraph{Error-correction probe (\Cref{fig:perturbation}).}
We use $S_3$ with one layer and each model's grid-best checkpoint. For
each of $N=200$ fresh sequences of length $T=200$ we run a clean rollout
$h^{\mathrm{clean}}_{i,t}$ and a perturbed rollout
$h^{\mathrm{pert}}_{i,t}$ that is identical up to step $t_0=20$, where we
inject i.i.d.\ Gaussian noise of standard deviation $\sigma=10^{-2}$ into
the recurrent state of the first layer (the state
that the recurrence propagates, not the residual stream). We then track
$e_{i,t}=h^{\mathrm{pert}}_{i,t}-h^{\mathrm{clean}}_{i,t}$ for $t\geq t_0$
and report the median of $\|e_{i,t}\|_2/\|e_{i,t_0}\|_2$. The per-step
contraction rate $\rho_{\mathrm{step}}$ quoted in the main text is the
$(T-t_0-1)$-th root of the median final ratio. The 3-D trajectory panels
project the error onto its leading two PCA components computed at $t_0$.

\paragraph{State separation over rollouts (\Cref{fig:codebook_collapse}).}
On $S_3$ with one layer and the grid-best checkpoints, we evaluate $N=200$ fresh sequences of length
$T_{\max}=1500$, well beyond $L_{\max}=60$. At every step $t$ we group
the rollouts by their oracle symbol and take per-class means to obtain
a time-resolved centroid $c_g(t)=\mathbb{E}_i[h_{i,t}\!\mid\!g_{i,t}=g]$.
The reported quantities are $R(t), M(t), q(t)$ from
\Cref{subsec:theory_error_budget}, instantiated as
$R(t)=\mathbb{E}_i\|W_{\mathrm{out}}(h_{i,t}-c_{g_{i,t}}(t))\|_2$ and
$M(t)=\min_{g\neq g'}\|W_{\mathrm{out}}(c_g(t)-c_{g'}(t))\|_2$. The
figure shows readout-space versions as the primary curves and the
latent-space versions (without $W_{\mathrm{out}}$) as gray overlays.
Vertical markers reproduce each model's max-passing length from
\Cref{tab:main}.

\paragraph{Crossing time (\Cref{fig:corr}).}
\label{ax:tcross}
To ignore brief spikes, we compute $q(t)$ from $50$-step moving averages of $R$ and $M$,
and take $T_{\mathrm{cross}}$ as the first step after the training length from which
$q(t)\geq 0.5$ holds for $50$ steps. The median $q(t)$ at $\mathrm{mp}$ in
\Cref{subsec:exp_analysis} uses the same averages; \Cref{fig:corr} plots the
unsmoothed ratio.

\paragraph{Symbolic-subspace decomposition (\Cref{fig:subspace_decomp}).}
We use the rollouts and checkpoints of \Cref{fig:codebook_collapse}. At each step
$t$ we form the centered centroid matrix $\widetilde C(t)\in
\mathbb{R}^{|G|\times d_{\mathrm{model}}}$ and take the top $k=|G|-1$
right singular vectors as the orthonormal basis $P_{\mathcal U}(t)$ of
the symbolic subspace. The within-class deviation
$\delta_{i,t}=h_{i,t}-c_{g_{i,t}}(t)$ splits into its squared component in the symbolic subspace,
$\|P_{\mathcal U}(t)^{\!\top}\delta_{i,t}\|^2$, and the orthogonal remainder,
$\|\delta_{i,t}\|^2-\|P_{\mathcal U}(t)^{\!\top}\delta_{i,t}\|^2$. We aggregate
each part across rollouts as a root mean square and normalize by the latent
inter-centroid scale
$r_{\mathrm{sep}}(t)=\min_{g\neq g'}\|c_g(t)-c_{g'}(t)\|_2$.

\paragraph{Nonlinear activation type (\Cref{tab:nonlinearities}).}
We fix the recurrence $h_t=\phi(W_hh_{t-1}+W_xx_t+b)$ on $S_3$ with one layer and vary only $\phi$: an affine baseline (identity),
two whole-vector normalizations (LayerNorm, sphere projection
$h\mapsto h/\|h\|$), pointwise nonlinearities ($\tanh$, ReLU), pointwise
pair operators ($\max,\min$), and GroupSort with group size
$k=2$ \citep{anil2019sorting}. The identity and $\tanh$ entries are the Linear RNN and tanh RNN of \Cref{tab:main}, with the full grid; each other operator uses one configuration ($d_{\mathrm{state}}=64$, learning rate $10^{-3}$, cosine schedule) with three seeds. Each cell reports the best max-passing length; \Cref{ax:nonlinearity_jacobians} analyzes the per-operator Jacobians.

\section{Proofs}
\label{ax:proof}

\subsection{Proof of \texorpdfstring{\Cref{thm:affine_neutrality}}{Theorem 1}}
\label{proof:affine_neutrality}

Let $s$ be a state-preserving sequence and assume the induced return map
is affine, $F_s(h) = A_s h + b_s$. By the exact-preservation hypothesis
of \Cref{thm:affine_neutrality}, $F_s(\cs{g}) = \cs{g}$ for every
$g \in G$. For any pair $g, g' \in G$,
\[
A_s(\cs{g} - \cs{g'})
= F_s(\cs{g}) - F_s(\cs{g'})
= \cs{g} - \cs{g'}.
\]
The vectors $\{\cs{g} - \cs{g'}\}_{g, g' \in G}$ span $\mathcal{U}$ by
\eqref{eq:symbolic_subspace}, so $A_s$ acts as the identity on
$\mathcal{U}$, establishing $A_s|_{\mathcal{U}} = I$. \qed

\subsection{Proof of perturbation neutrality}
\label{proof:no_symbolic_correction}

We verify the consequence stated in \Cref{subsec:theory_impossibility}:
under the hypotheses of \Cref{thm:affine_neutrality}, every
perturbation $\delta \in \mathcal{U}$ is transported unchanged by the
return map. Continuing under the affine return assumption, let
$g \in G$ and $\delta \in \mathcal{U}$. Then
\[
F_s(\cs{g} + \delta) - F_s(\cs{g})
= A_s(\cs{g} + \delta) + b_s - (A_s \cs{g} + b_s)
= A_s \delta
= \delta,
\]
where the last equality uses $A_s|_{\mathcal{U}} = I$ from
\Cref{thm:affine_neutrality}. Thus the perturbation along
$\mathcal{U}$ is preserved exactly under the return map. \qed

\subsection{Extension to non-minimal realizations}
\label{proof:nonminimal}

\Cref{thm:affine_neutrality} assumes one reference representation
$c^\star_g$ per symbolic state. A trained network may instead realize a
symbolic state by several hidden states, all decoded to the same symbol, which
may move or cycle among themselves under the dynamics, so that $F_s(h)=h$ need
not hold for any of them. Fix a state-preserving sequence $s$ and, for each
$g\in G$, let $\mu_g$ be a distribution over the hidden states that represent
$g$. Suppose the return map preserves it, $(F_s)_\sharp\mu_g=\mu_g$, and let
$m_g:=\mathbb E_{h\sim\mu_g}[h]$ be its centroid. Because $F_s(h)=A_s h+b_s$ is
affine,
\[
F_s(m_g)=F_s\bigl(\mathbb E_{\mu_g}[h]\bigr)=\mathbb E_{\mu_g}[F_s(h)]
=\mathbb E_{(F_s)_\sharp\mu_g}[h]=m_g .
\]
Hence, for any $g,g'\in G$,
$A_s(m_g-m_{g'})=F_s(m_g)-F_s(m_{g'})=m_g-m_{g'}$, so $A_s u=u$ for every
$u\in\mathcal U_{\mathrm{nm}}:=\operatorname{span}\{m_g-m_{g'}:g,g'\in G\}$.
The conclusion of \Cref{thm:affine_neutrality} therefore holds on the
subspace separating the preserved class centroids, provided $m_g\neq m_{g'}$
for $g\neq g'$. This covers the recurrent part of a finite non-minimal
realization, in which several hidden states represent the same symbolic state
and move within its decoder class while the class centroid stays fixed.

\paragraph{Limitation.}
The argument does not cover arbitrary decoder equivalence classes. If two
symbolic states have the same centroid, $m_g=m_{g'}$, their centroid
difference does not separate them and the argument says nothing about the
directions that do. Transient hidden states are not covered until they enter a
preserved class. The theorem therefore directly covers point-valued
representations $c^\star_g$ and extends to non-minimal realizations with
distinct preserved class centroids.

\subsection{Proof of \texorpdfstring{\Cref{cor:error_budget_modes}}{Corollary 1}}
\label{app:error_budget}

\paragraph{Affine neutrality on the symbolic subspace.}
Let \(F_s(h)=A_s h+b_s\). Since \(F_s\) is the exact
state-preserving affine return map considered in
\Cref{thm:affine_neutrality}, its linear part satisfies
\[
A_s|_{\mathcal U}=I .
\]
Thus, along directions that distinguish symbolic states, the exact affine
return map has no contracting homogeneous component. Along a return cycle,
\(g_{t+1}=g_t\), and exact state preservation gives
\(F_s(\cs{g_t})=\cs{g_t}\).

\paragraph{Projected error recurrence.}
Using \(\tilde F_s=F_s+\varepsilon\), the trained update satisfies
\[
h_{t+1}
=
\tilde F_s(h_t)
=
F_s(h_t)+\varepsilon(h_t).
\]
The projected deviation from the fixed reference representation \(c^\star_{g_t}\) of the current symbolic state (unchanged along the return cycle, since \(g_{t+1}=g_t\)) is therefore
\[
\begin{aligned}
e_{\mathcal U}(t+1)
&=
P_{\mathcal U}(h_{t+1}-\cs{g_t}) \\
&=
P_{\mathcal U}\bigl(F_s(h_t)-F_s(\cs{g_t})\bigr)
+
P_{\mathcal U}\varepsilon(h_t).
\end{aligned}
\]
The first term is the effect of the exact affine return map on the current
deviation, while the second term is the projected approximation residual of
the trained tracker. Since \(F_s\) is affine,
\[
F_s(h_t)-F_s(\cs{g_t})
=
A_s(h_t-\cs{g_t}).
\]
Restricting to the projected dynamics inside \(\mathcal U\), and using
\(A_s|_{\mathcal U}=I\), the exact affine part preserves the current
symbolic-subspace deviation rather than reducing it. Hence
\[
e_{\mathcal U}(t+1)
=
e_{\mathcal U}(t)+\eta_t,
\qquad
\eta_t:=P_{\mathcal U}\varepsilon(h_t).
\]
Unrolling this recurrence gives
\[
e_{\mathcal U}(t)
=
e_{\mathcal U}(0)+\sum_{j=0}^{t-1}\eta_j .
\]
Thus projected residuals are accumulated, not corrected, by the affine
return dynamics.
Here \(e_{\mathcal U}(0)\) is the error in the hidden state handed to the
recurrence, for example by an encoder or from a compressed prefix, and is
separate from the transition residual \(\varepsilon\); exact affine return
dynamics preserve it, \(e_{\mathcal U}(t)=e_{\mathcal U}(0)\).

\paragraph{Linear accumulation under coherent residual drift.}
The previous identity does not require residuals to grow linearly; it only
shows that they enter additively. Linear growth occurs when the residuals
have a coherent component over the relevant horizon. If
\[
\frac{1}{t}\sum_{j<t}\eta_j\approx \bar\eta\neq0,
\]
then
\[
\sum_{j<t}\eta_j\approx t\bar\eta,
\qquad
e_{\mathcal U}(t)\approx e_{\mathcal U}(0)+t\bar\eta .
\]
The fixed initial error is lower order relative to the growing drift, so the
readout-visible component has scale
\[
R(t)
\approx
\|W_{\mathrm{out}}(t\bar\eta)\|
=
t\|W_{\mathrm{out}}\bar\eta\|.
\]

\paragraph{Crossing scale.}
If the between-state separation remains approximately stable,
\(M(t)\approx M>0\), then
\[
q(t)=\frac{R(t)}{M(t)}
\approx
\frac{t\|W_{\mathrm{out}}\bar\eta\|}{M}.
\]
The crossing time is obtained by solving \(q(t)=\tau\), which yields
\[
T_{\mathrm{cross}}
\approx
\frac{\tau M}{\|W_{\mathrm{out}}\bar\eta\|}.
\]
This shows that the finite horizon is controlled by the competition between
the stable separation scale \(M\) and the rate of coherent readout-visible
drift \(\|W_{\mathrm{out}}\bar\eta\|\).
\qed

\section{Further \texorpdfstring{discussion}{discussion}}
\label{ax:further_discussions}

\subsection{Per-operator Jacobian analysis}
\label{ax:nonlinearity_jacobians}

\Cref{subsec:theory_impossibility} identifies a sufficient condition
for state-dependent error correction: the return-map Jacobian on
$\mathcal{U}$ has norm strictly below one uniformly over the references $\cs{g}$. The
activation $\phi$ in the canonical form is what gives a return map
state-dependent Jacobians at all, so the operational question is which
choices of $\phi$ can deliver this contraction.
\Cref{tab:nonlinearities} reports that pointwise activations and pair
operators support state tracking on $S_3$ while whole-vector
normalizations do not, despite both classes being nonlinear. The
distinction is visible in the local Jacobian of each operator. Writing
the recurrence as $h_t=\phi(p_t)$ with pre-activation
$p_t = W_h h_{t-1} + W_x x_t + b$, the Jacobian wrt $h_{t-1}$ at a fixed reference
$\cs{g}$ is
\[
J_\phi(\cs{g}) = \frac{\partial\phi}{\partial p}\Big|_{p_t}\,W_h.
\]
Whether the resulting linear map can act differently across different
references $\cs{g}, \cs{g'}$, and in particular contract along the
symbolic subspace $\mathcal{U}=\mathrm{span}\{\cs{g}-\cs{g'}\}$, depends
entirely on $\partial\phi/\partial p$.

\paragraph{Pointwise scalar activations: state-dependent diagonal.}
For $\tanh$, $\partial\phi/\partial p = \mathrm{diag}(1-\tanh^2(p_t))$,
which depends elementwise on $p_t$ and therefore on $h_{t-1}$. ReLU is the
binary case $\partial\phi/\partial p = \mathrm{diag}(\mathbf{1}[p_t>0])$.
In both cases the diagonal can scale entries below $1$ at some
references and not others, so the return-map Jacobian on $\mathcal{U}$ can
contract, exactly the conditional restoration that
\Cref{thm:affine_neutrality} forbids in the affine case.

\paragraph{Max and min: state-dependent masks.}
$\max$ and $\min$ take, in each coordinate, either the recurrent term
$W_h h_{t-1} + b_h$ or the input term $W_x x_t + b$, so
\[
J_{\max}(h_{t-1}) = \mathrm{diag}(s_t)\,W_h,
\]
where $s_t\in\{0,1\}^d$ marks the coordinates in which the recurrent term
is selected ($\min$ is analogous). The mask depends on $h_{t-1}$, so the
return-map Jacobian on $\mathcal{U}$ can contract at some references and
not at others.

\paragraph{GroupSort: piecewise permutations.}
\texttt{GroupSort} with $k=2$ sorts each pair of entries of $p_t$. Its
Jacobian is a \emph{permutation matrix} that varies with $h_{t-1}$, so the
composed return map mixes $\mathcal{U}$ directions differently at
different references. Because the permutation is also $1$-Lipschitz in
each linear region, the model can implement strict contraction without
the gain saturating.

\paragraph{Whole-vector normalizations: nearly identity on $\mathcal{U}$.}
For LayerNorm with mean $\mu$ and variance $\sigma^2$,
\[
\frac{\partial\phi}{\partial p}\Big|_{p_t} = \frac{1}{\sigma}\Bigl(I - \tfrac{1}{d}\mathbf{1}\mathbf{1}^{\!\top}
                                     - \tfrac{1}{d}\widetilde p_t\widetilde p_t^{\!\top}\Bigr),
\]
where $\widetilde p_t = (p_t-\mu)/\sigma$. The bracketed projector-like
operator (identity minus mean projection minus a rank-$1$ correction)
mixes $\mathcal U$ directions only through the rank-$1$ piece
$\widetilde p_t \widetilde p_t^{\!\top}/d$, whose contribution to a
generic direction in $\mathcal U$ is $O(1/d)$ and so cannot deliver a
fixed-margin contraction once $d \gg |G|$; the global prefactor
$1/\sigma$ is state-dependent but acts isotropically and cannot
discriminate between $\mathcal U$ directions. The unit-sphere
projection $h\mapsto h/\|h\|$ is similar: its Jacobian
$J(h) = (I-\hat h\hat h^{\!\top})/\|h\|$ is, up to the state-dependent
isotropic factor $1/\|h\|$, an orthogonal projector that does not
encode per-direction state information. We therefore expect both
operators to behave like affine maps on $\mathcal U$ at the per-state
level, leaving the obstruction of \Cref{thm:affine_neutrality} intact;
this is an intuition rather than a formal lemma.

In short, success on the nonlinearity probe depends on whether
$\partial\phi/\partial p$ encodes \emph{per-direction} state
information. Pointwise and pair operators do; norm operators do not.

\subsection{The \texorpdfstring{$C_2$}{C2} edge case}
\label{app:parity_edge_case}

\Cref{tab:main} shows that parity (\(C_2\)) is the only group on which
several affine models, notably Negative Mamba and Token-gated RNN, reach
the maximum tested length. This success does not imply genuine correction
of symbolic-subspace error. Instead, \(C_2\) is unusually tolerant of neutral
oscillation: errors can persist and flip sign while remaining within the
binary readout margin.

\paragraph{Neutral oscillation in \(C_2\).}
Let \(C_2=\{e,a\}\) with fixed references \(\cs{e},\cs{a}\in\mathbb{F}^d\),
\(\cs{e}\neq \cs{a}\), and define
\[
F_a(h)=-h+(\cs{e}+\cs{a}),
\qquad
F_e(h)=h .
\]
Then \(F_a(\cs{e})=\cs{a}\), \(F_a(\cs{a})=\cs{e}\), and
\(F_a^2=\mathrm{id}\), so the parity transition is realized exactly by an
affine involution. For a perturbation \(\delta\in\mathcal U\),
\[
F_a(\cs{g}+\delta)=\cs{g\cdot a}-\delta,
\qquad
F_a^2(\cs{g}+\delta)=\cs{g}+\delta .
\]%
Thus the perturbation is transported as an oscillation rather than
contracted. The prediction can nevertheless remain correct as long as the
oscillation stays inside the nearest-centroid margin.

\paragraph{Tolerance shrinks with cycle order.}
This margin advantage is largest for \(C_2\). In the canonical regular
\(C_k\) geometry, with
\[
\cs{i}=r(\cos(2\pi i/k),\sin(2\pi i/k)),
\]
adjacent references are separated by
\[
\|\cs{i+1}-\cs{i}\|=2r\sin(\pi/k).
\]%
Nearest-centroid decoding therefore tolerates only a perturbation on the
order of
\[
r\sin(\pi/k)\approx \frac{\pi r}{k}
\]
before confusing neighboring states. Equivalently, the angular decision
sector has width \(2\pi/k\), so the tolerated phase error is only
\(\pi/k\). Thus the neutral oscillation that can be harmless for \(C_2\)
becomes increasingly fragile as the cycle order grows.

\paragraph{Relation to affine neutrality.}
This behavior is consistent with \Cref{thm:affine_neutrality}. A
state-preserving affine return map acts as the identity on
\(\mathcal U\), so it cannot contract symbolic-subspace perturbations. In
\(C_2\), the binary margin can hide this lack of contraction over the
tested horizons. For larger cycles, the margin is smaller and the return map must remain tightly aligned with the cyclic geometry of the references;
residual phase or orbit mismatch is transported rather than corrected.

\paragraph{Return-\texorpdfstring{map}{map} gain diagnostic.}
A practical diagnostic is the spectral radius of \(A_s|_{\mathcal U}\)
over short state-preserving sequences \(s\). Gain near \(1\), together
with non-degenerate centroids, indicates that errors persist without being
contracted. Gain \(>1\) indicates amplification, while gain \(<1\) can
reflect either centroid collapse or genuinely state-dependent correction.
On \(C_2\), Negative Mamba and Token-gated RNN have median return-map gain
near \(1\), consistent with the neutral oscillatory route.

\subsection{Effect of task complexity}
\label{app:task_complexity}

The number of symbolic states enters \Cref{cor:error_budget_modes} only
implicitly, through the minimum readout-space separation $M$. To make this
dependence explicit, fix a step $t$ in the regime $M(t)\approx M$, and let
$z_g=W_{\mathrm{out}}c_g(t)$, $\bar z=|G|^{-1}\sum_g z_g$,
$r=\dim\mathrm{span}\{z_g-\bar z\}\geq 1$, and
$B=\max_g\|z_g-\bar z\|_2$. With $M=\min_{g\neq g'}\|z_g-z_{g'}\|_2$, the
balls of radius $M/2$ around the $z_g$ are disjoint and lie in the ball of
radius $B+M/2$ around $\bar z$ within the $r$-dimensional affine hull of the
$z_g$. Comparing volumes gives $|G|(M/2)^r\leq(B+M/2)^r$, and therefore
\[
M\leq\frac{2B}{|G|^{1/r}-1}.
\]%
Hence, under a fixed effective rank $r$ and a bounded readout-space radius
$B$, the upper bound on the achievable minimum separation decreases as the
number of symbolic states grows. Since \Cref{cor:error_budget_modes} predicts
$T_{\mathrm{cross}}\approx\tau M/\|W_{\mathrm{out}}\bar\eta\|$, this
correspondingly lowers the best separation-limited crossing scale when the
residual drift is held approximately fixed.

This does not make $T_{\mathrm{cross}}$ monotone in $|G|$: a model can raise
its effective rank or scale, and harder transitions can change $\bar\eta$.
In the corollary, $|G|$ enters only through $M$.

\section{Additional empirical results}
\label{ax:more_results}

\subsection{Additional models}
\label{subsec:ax_more_models}

\Cref{tab:more_models} extends \Cref{tab:main} to four additional
architectures absent from the main table:
PD-SSM~\citep{terzic2025structured},
DeltaNet~\citep{yang2024deltanet},
DeltaProduct~\citep{siems2025deltaproduct}, and Low-Rank RNN, a tanh RNN whose recurrent matrix has rank $2$, alongside the models from
\Cref{tab:main}.

\begin{table}[t]
    \centering
    \small
    \setlength{\tabcolsep}{4pt}
    \begin{tabular}{llcccccc}
        \toprule
        Family & Model & \multicolumn{2}{c}{$C_2$} & \multicolumn{2}{c}{$C_6$} & \multicolumn{2}{c}{$S_3$} \\
        \cmidrule(lr){3-4} \cmidrule(lr){5-6} \cmidrule(lr){7-8}
         &  & L1 & L2 & L1 & L2 & L1 & L2 \\
        \midrule
        \multirow{2}{*}{Diagonal SSM} & Mamba & \bad & 60 & \bad & 60 & \bad & \bad \\
         & Negative Mamba & 1000 & 1000 & 100 & 200 & 100 & 200 \\
        \midrule
        \multirow{3}{*}{Complex SSM} & Mamba-3 & 200 & 300 & 100 & 100 & \bad & 60 \\
         & AUSSM & 1000 & \bad & 200 & 100 & \bad & \bad \\
         & Simple AUSSM & 300 & 400 & 100 & 100 & 60 & 100 \\
        \midrule
        \multirow{1}{*}{Sparse SSM} & PD-SSM & 300 & 600 & 400 & 300 & 100 & 60 \\
        \midrule
        \multirow{3}{*}{Delta-rule} & DeltaNet & \bad & \bad & \bad & 60 & \bad & \bad \\
         & DeltaProduct & \bad & \bad & \bad & 60 & \bad & 100 \\
        \midrule
        \multirow{1}{*}{Linear RNN} & Linear RNN & \bad & 100 & \bad & 60 & \bad & \bad \\
        \midrule
        \multirow{1}{*}{Input-gated RNN} & Token-gated RNN & 1000 & 700 & 300 & 400 & 500 & 1000 \\
        \midrule
        \multirow{3}{*}{State-dependent} & tanh RNN & 1000 & 1000 & 1000 & 1000 & 1000 & 1000 \\
         & Low-Rank RNN ($r{=}2$, tanh) & 1000 & 1000 & \bad & 1000 & 1000 & 1000 \\
         & State-gated RNN & 1000 & 1000 & 1000 & 1000 & 1000 & 1000 \\
        \bottomrule
    \end{tabular}
    \vspace{0.5em}
    \caption{\textbf{Max-passing length of additional models.}
We extend~\Cref{tab:main} to additional existing architectures:
PD-SSM~\citep{terzic2025structured}, DeltaNet~\citep{yang2024deltanet}, DeltaProduct~\citep{siems2025deltaproduct}, and a rank-$2$ tanh RNN.
The results show that learned solutions need not realize the full expressivity
available to the architecture.
}
    \label{tab:more_models}
\end{table}

\subsection{Delta-rule models}
\label{subsec:ax_delta_rule}

The delta-rule models in \Cref{tab:main} update a matrix-valued state whose transition depends only on the input, so they are affine in the state and \Cref{thm:affine_neutrality} applies to them (\Cref{ax:model}). They allow negative eigenvalues~\citep{grazzi2024unlocking}.

\begin{figure}[t]
    \centering
    \includegraphics[width=0.98\linewidth]{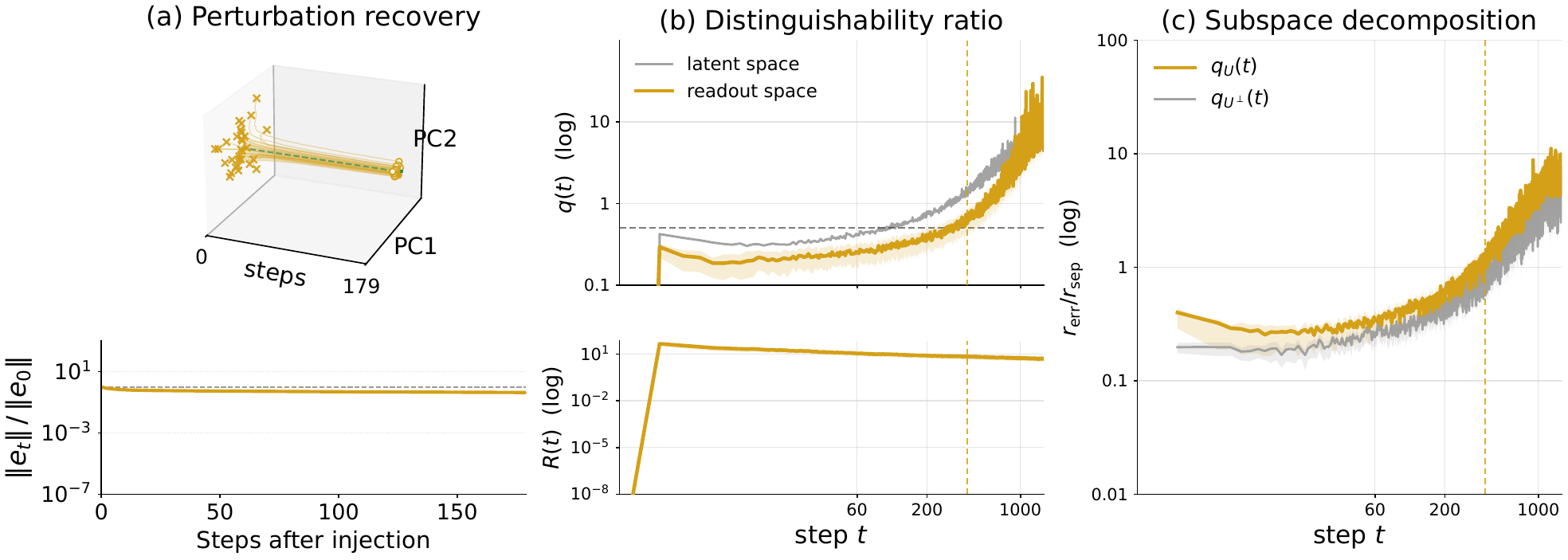}
    \caption{\textbf{Error control of Gated DeltaProduct on $S_3$.} We repeat the analyses of Figures~\labelcref{fig:perturbation}, \labelcref{fig:codebook_collapse}, and~\labelcref{fig:subspace_decomp}. Vertical dashes mark its max-passing length of $400$. The injected error is not removed, and $q(t)$ climbs past the nearest-centroid bound before that length.}
    \label{fig:delta_rule}
\end{figure}

Of the four models, only Gated DeltaProduct passes the curriculum on $S_3$ with a single layer. Hence, as shown in \Cref{fig:delta_rule}, we focus our analysis on this model, replicating the results of Figures~\labelcref{fig:perturbation}, \labelcref{fig:codebook_collapse}, and~\labelcref{fig:subspace_decomp}. Like the other affine models, it does not remove the injected error: the median error ratio only falls to $0.44$ (\Cref{fig:delta_rule}a), whereas the state-dependent models reduce it by several orders of magnitude (\Cref{fig:perturbation}). Its distinguishability ratio $q(t)$ follows the climb regime of \Cref{subsec:theory_error_budget} (\Cref{fig:delta_rule}b). It starts below the nearest-centroid bound, where states are still closer to their own centroid than to any other, and climbs past the bound before the max-passing length as the spread grows relative to the class separation. In the subspace decomposition, the spread along the symbolic subspace $\mathcal U$ is larger than the orthogonal spread from the start (\Cref{fig:delta_rule}c), so the growing error already lies in the directions that separate the states, where affine dynamics cannot contract it (\Cref{thm:affine_neutrality}).

These results also suggest why Gated DeltaProduct extrapolates further than the other affine models that pass $S_3$. As an affine model, it cannot reduce errors along $\mathcal U$ without also reducing the separation between states. It does not escape this trade-off, but it settles at a better point on it: it starts extrapolation with a spread that is small relative to the separation, so the accumulated spread takes longer to reach the bound, as \Cref{cor:error_budget_modes} predicts. The longer horizon thus reflects a better compromise between separation and error reduction rather than error correction.

\subsection{Additional tasks}
\label{subsec:ax_more_tasks}

\Cref{tab:more_tasks} extends \Cref{tab:main} to two additional groups:
the Abelian group \(C_2\times C_4\) and the non-Abelian alternating
group \(A_4\). Both are evaluated on the five recurrent architectures
that exhibit non-trivial behavior in \Cref{tab:main}: tanh
RNN~\citep{elman1990finding}, State-gated RNN, Token-gated RNN,
Negative Mamba, and Mamba-3~\citep{lahoti2026mamba3}.

\begin{table}[t]
    \centering
    \small
    \setlength{\tabcolsep}{6pt}
    \begin{tabular}{l cc cc}
        \toprule
        & \multicolumn{2}{c}{$C_2 \times C_4$} & \multicolumn{2}{c}{$A_4$} \\
        \cmidrule(lr){2-3} \cmidrule(lr){4-5}
        Model & L1 & L2 & L1 & L2 \\
        \midrule
        tanh RNN          & 1000 & 1000 & 1000 & 1000 \\
        State-gated RNN   & 1000 & 1000 & 1000 & 1000 \\
        \midrule
        Token-gated RNN   &  200 &  300 &  500 &  300 \\
        Negative Mamba    &   60 &  300 & \bad & \bad \\
        Mamba-3            & \bad &  100 & \bad & \bad \\
        \bottomrule
    \end{tabular}
    \vspace{0.5em}
    \caption{\textbf{Model performances on additional tasks.} We extend~\Cref{tab:main} to $C_2 \times C_4$ and $A_4$. Both tasks exhibit trends consistent with the main results.}
    \label{tab:more_tasks}
\end{table}

\subsection{\texorpdfstring{Per-architecture spread at a fixed step}{Per-architecture spread at a fixed step}}
\label{subsec:ax_stat}

\Cref{fig:subspace_decomp} displays per-step root-mean-square spreads
and \Cref{fig:perturbation} per-step medians, both with IQR bands over
$N=200$ rollouts. \Cref{tab:per_arch_summary}
tabulates median, $Q_1$--$Q_3$, and max over the same rollouts at a
single representative step $t_\mathrm{eval}$, on the same grid-best
checkpoints used in the figures. The metric definitions
match the figures: per-rollout latent $\|P_{\mathcal U}\delta\|$ divided by the
minimum class-pair separation for
$r_{\mathrm{err},\mathcal U}/r_{\mathrm{sep}}$; unprojected
per-rollout error norm for $\|e\|/\|e_{t_0}\|$. We pick
$t_\mathrm{eval}$ as the max-passing length when the model learned the task and
the curriculum length $60$ otherwise, since the trajectories of failed models diverge numerically beyond that point.

\begin{table}[t]
  \centering
  \footnotesize
  \setlength{\tabcolsep}{1.8pt}
  \begin{tabular}{l r r cc}
    \toprule
    Model & \begin{tabular}[b]{@{}r@{}}Max-passing\\length\end{tabular} & $t_\mathrm{eval}$ & $r_{\mathrm{err},\mathcal U}/r_{\mathrm{sep}}$ (latent) & $\|e\|/\|e_{t_0}\|$ \\
    \midrule
    Mamba & \bad & 60 & $1.69$ [$1.20$, $2.27$] / $8.21$ & $0.26$ [$0.25$, $0.26$] / $0.26$ \\
    Mamba-3 & \bad & 60 & $9.62$ [$7.04$, $11.92$] / $19.32$ & $0.94$ [$0.94$, $0.94$] / $0.95$ \\
    Negative Mamba & 100 & 100 & $3.55$ [$2.79$, $6.03$] / $26.29$ & $0.47$ [$0.43$, $0.50$] / $0.64$ \\
    Token-gated RNN & 500 & 500 & $1.39$ [$1.00$, $2.00$] / $7.31$ & $4.0\times10^{11}$ [$2.6\times10^{11}$, $6.4\times10^{11}$] / $2.4\times10^{12}$ \\
    tanh RNN & 1000 & 1000 & $0.65$ [$0.55$, $0.83$] / $1.27$ & $4.9\times10^{-6}$ [$4.6\times10^{-6}$, $5.2\times10^{-6}$] / $6.5\times10^{-6}$ \\
    State-gated RNN & 1000 & 1000 & $0.67$ [$0.52$, $0.87$] / $1.65$ & $3.9\times10^{-5}$ [$3.5\times10^{-5}$, $4.4\times10^{-5}$] / $6.2\times10^{-5}$ \\
    \bottomrule
  \end{tabular}
  \vspace{0.5em}
  \caption{\textbf{Per-architecture error bars for \Cref{fig:subspace_decomp} and \Cref{fig:perturbation} at $t_\mathrm{eval}$ on $S_3$ with one layer, grid-best checkpoints.}
  Each cell shows median $[Q_1, Q_3]$ / max over $N=200$ rollouts. $t_\mathrm{eval}$ is the model's max-passing length when it learned the task, or the curriculum length $60$ when it did not.
  $r_{\mathrm{err},\mathcal U}/r_{\mathrm{sep}}$: per-rollout $\|P_{\mathcal U}\delta\|$ divided by the minimum class-pair separation (latent).
  $\|e\|/\|e_{t_0}\|$: per-rollout perturbation-error ratio from the probe of \Cref{fig:perturbation}, at step $t_\mathrm{eval}$.}
  \label{tab:per_arch_summary}
\end{table}

\subsection{Seed variability}
\label{subsec:ax_seeds}

\Cref{tab:main} reports the best max-passing length across seeds, following
\citet{shakerinava2026expressive}; \Cref{tab:seed_variability} reports the
mean and standard deviation across seeds. The state-dependent models reach
$1000$ on every seed in all six settings, whereas most affine models that
extrapolate beyond the training length vary across seeds.

\begin{table}[t]
  \centering
  \footnotesize
  \setlength{\tabcolsep}{4pt}
  \begin{tabular}{l cccccc}
    \toprule
    Model & $C_{2}$ L1 & $C_{2}$ L2 & $C_{6}$ L1 & $C_{6}$ L2 & $S_{3}$ L1 & $S_{3}$ L2 \\
    \midrule
    Mamba & $0\pm0$ & $60\pm0$ & $0\pm0$ & $60\pm0$ & $0\pm0$ & $0\pm0$ \\
    Mamba-3 & $167\pm58$ & $140\pm139$ & $100\pm0$ & $100\pm0$ & $0\pm0$ & $20\pm35$ \\
    AUSSM & $533\pm451$ & $0\pm0$ & $133\pm58$ & $67\pm58$ & $0\pm0$ & $0\pm0$ \\
    Simple AUSSM & $500\pm436$ & $187\pm186$ & $100\pm0$ & $87\pm23$ & $40\pm35$ & $87\pm23$ \\
    Negative Mamba & $1000\pm0$ & $1000\pm0$ & $100\pm0$ & $133\pm58$ & $100\pm0$ & $133\pm58$ \\
    Linear RNN & $0\pm0$ & $100\pm0$ & $0\pm0$ & $20\pm35$ & $0\pm0$ & $0\pm0$ \\
    Token-gated RNN & $1000\pm0$ & $300\pm346$ & $233\pm58$ & $200\pm173$ & $367\pm115$ & $467\pm462$ \\
    \midrule
    tanh RNN & $1000\pm0$ & $1000\pm0$ & $1000\pm0$ & $1000\pm0$ & $1000\pm0$ & $1000\pm0$ \\
    State-gated RNN & $1000\pm0$ & $1000\pm0$ & $1000\pm0$ & $1000\pm0$ & $1000\pm0$ & $1000\pm0$ \\
    \bottomrule
  \end{tabular}
  \vspace{0.5em}
  \caption{\textbf{Mean $\pm$ standard deviation of the max-passing length over three seeds.} Each entry uses the hyperparameter setting (state dimension, learning rate and scheduler) behind the value in \Cref{tab:main}; when several settings reach it, the one with the highest mean. Failure (\bad) counts as $0$.}
  \label{tab:seed_variability}
\end{table}

\subsection{Sensitivity of \texorpdfstring{$T_{\mathrm{cross}}$}{T\_cross} to the threshold and the measurement space}
\label{subsec:ax_q_sensitivity}

\Cref{tab:q_sensitivity} recomputes $T_{\mathrm{cross}}$ for other
thresholds $\tau$ and in latent space, without $W_{\mathrm{out}}$. In readout
space the correlation with the max-passing length barely changes
for $\tau$ from $0.5$ to $4$, so the result does not hinge on $\tau=1/2$.
Latent-space $q(t)$ is larger, so every run crosses $\tau\leq 1$ on the first step after the training length; at
thresholds matched to that scale the correlation returns. We use readout space because $\tau=1/2$ is
the nearest-centroid bound in the space the decoder reads.

\begin{table}[t]
  \centering
  \small
  \begin{tabular}{l l ccccc}
    \toprule
    Space & & $\tau=0.25$ & $\tau=0.5$ & $\tau=1$ & $\tau=2$ & $\tau=4$ \\
    \midrule
    Readout & $r$ & $+0.32$ & $+0.87$ & $+0.90$ & $+0.88$ & $+0.89$ \\
     & $n$ & 113 & 107 & 98 & 98 & 83 \\
    \midrule
    Latent & $r$ & -- & -- & -- & $+0.77$ & $+0.93$ \\
     & $n$ & 113 & 113 & 113 & 101 & 74 \\
    \bottomrule
  \end{tabular}
  \vspace{0.5em}
  \caption{\textbf{Sensitivity of $T_{\mathrm{cross}}$ to the threshold $\tau$ and the measurement space.} $r$ is the log-log Pearson correlation between $T_{\mathrm{cross}}$ and the max-passing length over the $n$ $S_3$ runs with $\mathrm{mp}\geq 60$ that cross. --: every run crosses on the first step after the training length, so $T_{\mathrm{cross}}$ is constant. $\tau=1/2$ in readout space is the setting of the main text.}
  \label{tab:q_sensitivity}
\end{table}

\subsection{Transformer baselines}
\label{subsec:ax_transformer}

Full-attention Transformers do not maintain a recurrent hidden state, so our
error analysis does not apply to them. For comparison, \Cref{tab:transformer}
reports the max-passing length of a causal Transformer with rotary position
embeddings and a learned beginning-of-sequence token, under the same tasks,
curriculum, depths and hyperparameter grid as \Cref{tab:main}. It reaches the
training length on $C_2$ and $C_6$ but does not extrapolate beyond it, and it
does not complete the curriculum on $S_3$, whereas both state-dependent models
reach $1000$ on all three tasks. On $S_3$ it fits the training sequences
without generalizing to held-out ones, consistent with reports that
Transformers need far more data than recurrent models to learn state
tracking~\citep{ebrahimi2026induction}.

\begin{table}[t]
    \centering
    \small
    \begin{tabular}{l cccccc}
        \toprule
        & \multicolumn{2}{c}{$C_{2}$} & \multicolumn{2}{c}{$C_{6}$} & \multicolumn{2}{c}{$S_{3}$} \\
        \cmidrule(lr){2-3} \cmidrule(lr){4-5} \cmidrule(lr){6-7}
        Model & L1 & L2 & L1 & L2 & L1 & L2 \\
        \midrule
        Transformer & 60 & 60 & 60 & 60 & \bad & \bad \\
        \bottomrule
    \end{tabular}
    \vspace{0.5em}
    \caption{\textbf{Max-passing length of a causal Transformer.} The model uses rotary position embeddings and a learned beginning-of-sequence token and is selected as in \Cref{tab:main}. $60$ denotes a model that passes only at the training length; \bad~denotes failure to complete the curriculum.}
    \label{tab:transformer}
\end{table}

\section{Examples of \texorpdfstring{state-tracking tasks}{state-tracking tasks}}
\label{ax:task}
\subsection{Parity (\texorpdfstring{$C_2$}{C2})}
The parity task is equivalent to modulo 2 counting, representing the state transitions within the cyclic group $C_2$. The algebraic structure of this group is detailed in the Cayley table (Table \ref{tab:cayley_c2}), while an example of parity tracking is provided in Example \ref{ex:c2}.
\begin{example}[\(C_2\)]
\label{ex:c2}
Let \(C_2=\{0,1\}\) with transition \(g_t = g_{t-1} + x_t \pmod 2\), where each input token \(x_t \in \{0,1\}\). Starting from \(g_0=0\), the input sequence \(1,1,0\) produces
\[
g_1 = 1\,\text{(Odd)},\qquad g_2 = 0\,\text{(Even)},\qquad g_3 = 0\,\text{(Even)}.
\]
The task is to output the running sum modulo \(2\) at every step.
\end{example}
\begin{table}[h]
\centering
\renewcommand{\arraystretch}{1.2}
\begin{tabular}{c|cc}
$\cdot$ & $0$ & $1$ \\ \hline
$0$ & $0$ & $1$ \\
$1$ & $1$ & $0$
\end{tabular}
\vspace{0.5em}
\caption{\textbf{Cayley table of $C_2$.}}
\label{tab:cayley_c2}
\end{table}
\subsubsection{Sketch of an \texorpdfstring{affine recurrent model for tracking}{affine recurrent model for tracking} \texorpdfstring{$C_2$}{C2}}

As demonstrated by \citet{sarrof2024expressive, grazzi2024unlocking}, solving the parity task with an affine recurrent model with real diagonal transitions, such as Mamba, requires the transition matrix to have at least one negative eigenvalue. To fulfill this condition and track the state of the $C_2$ group, we can design a minimalist 1-dimensional recurrent model as follows:
\begin{align}
h_t &= A(x_t) h_{t-1}, \quad h_0 = 1 \\
A(x_t) &= 
\begin{cases}
1 & \text{if } x_t = 0 \\
-1 & \text{if } x_t = 1
\end{cases}
\end{align}

Here, $h_t \in \{1, -1\}$ represents the internal state encoding the cumulative parity at time step $t$, initialized at $h_0 = 1$ corresponding to the identity element. The input-dependent transition parameter $A(x_t)$ applies the required negative eigenvalue ($-1$) whenever an active token ($x_t = 1$) is encountered. Consequently, each occurrence of $x_t=1$ inverts the sign of the hidden state, alternating between the two states of $C_2$. At the end of the sequence, the final state $h_T$ determines the parity: $h_T = 1$ indicates an even number of ones, whereas $h_T = -1$ indicates an odd number.

\subsection{Cyclic \texorpdfstring{group}{group} (\texorpdfstring{$C_3$}{C3})}
More generally, tracking operations within a cyclic group $C_k$ is equivalent to modulo $k$ counting, which tracks the cumulative sum of inputs wrapping around a finite set of $k$ distinct states in an Abelian (commutative) group. For the specific case of $C_3$, this corresponds to modulo 3 counting. The algebraic structure of $C_3$ is detailed in the Cayley table (Table \ref{tab:cayley_c3}), while an example of this state tracking is provided in Example \ref{ex:c3}.
\begin{table}[h]
\centering
\renewcommand{\arraystretch}{1.2}
\begin{tabular}{c|ccc}
$\cdot$ & $0$ & $1$ & $2$ \\ \hline
$0$ & $0$ & $1$ & $2$ \\
$1$ & $1$ & $2$ & $0$ \\
$2$ & $2$ & $0$ & $1$
\end{tabular}
\vspace{0.5em}
\caption{\textbf{Cayley table of $C_3$.}}
\label{tab:cayley_c3}
\end{table}

\begin{example}[\(C_3\)]
\label{ex:c3}
Let \(C_3=\{0,1,2\}\) with transition \(g_t = g_{t-1} + x_t \pmod 3\), where each input token \(x_t \in \{0,1,2\}\). Starting from \(g_0=0\), the input sequence \(1,2,1\) produces
\[
g_1 = 1,\qquad g_2 = 0,\qquad g_3 = 1.
\]
The task is to output the running sum modulo \(3\) at every step.
\end{example}

\subsubsection{Sketch of an \texorpdfstring{affine recurrent model for tracking}{affine recurrent model for tracking} \texorpdfstring{$C_3$}{C3}}

As proven by \citet{grazzi2024unlocking}, a linear recurrent model can count modulo $k$ (for non-power-of-two $k$, such as $k=3$) only if its transition matrix possesses at least one eigenvalue with a non-zero imaginary part ($\lambda \notin \mathbb{R}$). To fulfill this requirement and track the state of the cyclic group $C_3$, the model's capacity must be extended beyond the real number line to the complex domain. Specifically, we can formulate a minimalist 1-dimensional complex-valued recurrent model utilizing the 3rd roots of unity:
\begin{align}
h_t &= A(x_t) h_{t-1}, \quad h_0 = 1 + 0j \\
A(x_t) &= \exp\left(j \frac{2\pi}{3} x_t\right)
\end{align}
Here, $h_t \in \mathbb{C}$ represents the internal complex state at time step $t$, initialized at $h_0 = 1 + 0j$, which corresponds to the identity element (zero rotation). The input $x_t \in \{0, 1, 2\}$ indicates the degree of cyclic shift. The transition parameter $A(x_t)$ acts as a phase modulator, shifting the phase of the hidden state by exactly $120^\circ$ ($\frac{2\pi}{3}$ radians) multiplied by the input $x_t$. By operating on the unit circle in the complex plane, the model avoids exponential decay or explosion. At the end of the sequence, the final state $h_T$ encodes the modulo 3 sum of the inputs: $h_T = 1$ indicates $0$, $h_T = e^{j2\pi/3}$ indicates $1$, and $h_T = e^{j4\pi/3}$ indicates $2$.

\subsection{Symmetric \texorpdfstring{group}{group} (\texorpdfstring{$S_3$}{S3})}
The symmetric group $S_k$ comprises all possible permutations of a set containing $k$ distinct elements. The order of the group, representing the total number of permutations, is given by $\vert S_k\vert = k!$. While $S_1$ and $S_2$ are Abelian (commutative), $S_k$ exhibits non-commutative properties for all $k\geq 3$. Notably, $S_3$ is the smallest non-Abelian symmetric group, consisting of the six elements $\{e,(12),(23),(13),(123),(132)\}$. Example \ref{ex:non_abelian_s3} shows the non-Abelian nature of $S_3$.

In this set, $e$ denotes the identity element, which represents the permutation where all elements remain in their original positions. The cycle notation $(a\,b)$ represents a swapping that interchanges the positions of elements $a$ and $b$ while leaving the third element unchanged. In contrast, a 3-cycle such as $(a\,b\,c)$ denotes a cyclic permutation where the elements are shifted in a closed loop: $a$ moves to $b$, $b$ moves to $c$, and $c$ returns to $a$. The Cayley table of symmetric group $S_3$ is given in Table \ref{tab:cayley_s3}.

\begin{table}[h]
\centering
\renewcommand{\arraystretch}{1.2}
\begin{tabular}{c|cccccc}
$\cdot$ & $e$ & $(12)$ & $(13)$ & $(23)$ & $(123)$ & $(132)$ \\ \hline
$e$ & $e$ & $(12)$ & $(13)$ & $(23)$ & $(123)$ & $(132)$ \\
$(12)$ & $(12)$ & $e$ & $(132)$ & $(123)$ & $(23)$ & $(13)$ \\
$(13)$ & $(13)$ & $(123)$ & $e$ & $(132)$ & $(12)$ & $(23)$ \\
$(23)$ & $(23)$ & $(132)$ & $(123)$ & $e$ & $(13)$ & $(12)$ \\
$(123)$ & $(123)$ & $(13)$ & $(23)$ & $(12)$ & $(132)$ & $e$ \\
$(132)$ & $(132)$ & $(23)$ & $(12)$ & $(13)$ & $e$ & $(123)$
\end{tabular}
\vspace{0.5em}
\caption{\textbf{Cayley table of $S_3$.} The entry in row $a$ and column $b$ is $a\cdot b$, the permutation that applies $b$ first and then $a$.}
\label{tab:cayley_s3}
\end{table}

\begin{example}[Non-Abelian Property of $S_3$]
\label{ex:non_abelian_s3}
To illustrate the non-Abelian nature of \(S_3\), we compare two sequences with identical tokens in different orders. Let the input tokens be the generators \(t_1=(12)\) and \(t_2=(23)\).
\begin{enumerate}
\item Applying \((12)\) followed by \((23)\): \[g_0 = e \xrightarrow{(12)} g_1 = (12) \xrightarrow{(23)} g_2 = (123)\]
\item Applying \((23)\) followed by \((12)\): \[g'_0 = e \xrightarrow{(23)} g'_1 = (23) \xrightarrow{(12)} g'_2 = (132)\]
\end{enumerate}
Since the final states differ (\(g_2 \neq g'_2\)), the group \(S_3\) is non-Abelian.
\end{example}

\begin{table}[t]
    \centering
    \small
    \setlength{\tabcolsep}{8pt}
    \begin{tabular}{c c cc cc}
        \toprule
        & & \multicolumn{2}{c}{Input Indicators} & \multicolumn{2}{c}{Layer States} \\
        \cmidrule(lr){3-4} \cmidrule(lr){5-6}
        Input Token & Decomposition & $x^{(1)}_t$ ($C_2$) & $x^{(2)}_t$ ($C_3$) & $y^{(1)}_t$ & $h^{(2)}_t$ \\
        \midrule
        $e$     & $s^0 r^0$ & 0 & 0 &  1 & $1$ \\
        $(123)$ & $s^0 r^1$ & 0 & 1 &  1 & $e^{j 2\pi/3}$ \\
        $(132)$ & $s^0 r^2$ & 0 & 2 &  1 & $e^{j 4\pi/3}$ \\
        $(12)$  & $s^1 r^0$ & 1 & 0 & $-1$ & $1$ \\
        $(23)$  & $s^1 r^1$ & 1 & 1 & $-1$ & $e^{-j 2\pi/3}$ \\
        $(13)$  & $s^1 r^2$ & 1 & 2 & $-1$ & $e^{-j 4\pi/3}$ \\
        \bottomrule
    \end{tabular}
    \vspace{0.5em}
    \caption{\textbf{Mapping of $S_3$ elements.} The table demonstrates the bijective relationship between the input tokens, the decomposed indicators, and the internal model states $(y^{(1)}_t, h^{(2)}_t)$.}
    \label{tab:s3_mapping}
\end{table}

\subsubsection{Sketch of an \texorpdfstring{affine recurrent model for tracking}{affine recurrent model for tracking} \texorpdfstring{$S_3$}{S3}}
Here, we revisit the theoretical result of~\citet{shakerinava2026expressive}. Refer to the paper for a detailed analysis.
Recall that any operation in $S_3$ can be constructed using two fundamental generators: a transposition $s=(12)$, which swaps two elements, and a rotation $r=(123)$, which shifts the elements cyclically. These generators correspond to the parity group $C_2$ and the cyclic group $C_3$, respectively. To process these operations, we assume each input token at time $t$ can be decomposed into two corresponding attributes: a $C_2$ indicator $x^{(1)}_t$$\in\{0,1\}$ and a $C_3$ indicator $x^{(2)}_t$$\in\{0,1,2\}$. The model tracks the overall group state through the following layers:
\paragraph{Layer 1: $C_2$ parity tracker.}
The first layer operates as a 1-dimensional real-valued SSM that tracks the cumulative parity of the swapping operations, modeling the $C_2$ component.
\begin{itemize}
    \item \textbf{State transition:} Let $h_t^{(1)}\in\R$ be the hidden state initialized at $h_0^{(1)}=1$. The input-dependent transition parameter is defined as $A_t^{(1)}=(-1)^{x^{(1)}_t}$.
    \item \textbf{Update rule:} $h_t^{(1)}=A_t^{(1)}\cdot h_{t-1}^{(1)}$
    \item \textbf{Output:} $y_t^{(1)}=h_t^{(1)}\in\{1,-1\}$. This output indicates whether the current accumulated state is in a normal ($1$) or flipped ($-1$) orientation, capturing the $C_2$ state.
\end{itemize} 
\paragraph{Layer 2: conditional $C_3$ accumulator.}
The second layer operates as a 1-dimensional complex-valued SSM that tracks the rotational operations ($C_3$ component) in the complex plane. The phase shift in this layer is modulated by the output of Layer 1.
\begin{itemize}
    \item \textbf{State transition:} Let $h_t^{(2)}\in\C$ be initialized at $h_0^{(2)}=1$. The transition parameter $A_t^{(2)}$ is conditioned on the $C_2$ state $y_t^{(1)}$:
    \begin{equation}
        A_t^{(2)}=\exp\left(j\frac{2\pi}{3}\cdot x^{(2)}_t\cdot y_t^{(1)}\right)
    \end{equation}
    \item \textbf{Update rule:} $h_t^{(2)}=A_t^{(2)}\cdot h_{t-1}^{(2)}$
\end{itemize}

In summary, the internal layer states of the proposed 2-layer affine recurrent model track all six distinct elements of $S_3$. As demonstrated in Table \ref{tab:s3_mapping}, the unique combinations of these states provide a bijective mapping to the group elements without ambiguity. In practical implementations, although models do not explicitly partition these layers, residual connections and high-dimensional state spaces integrate these features, so the final layer can decode such non-commutative dynamics.
\section{Model \texorpdfstring{details}{details}}
\label{ax:model}

We map each architecture of \Cref{tab:main}, and related ones, to the canonical form
defined in \Cref{eq:canonical_state} and \Cref{eq:canonical_output}. For
each model we specify (i) the \textbf{canonical-form} realization, with
emphasis on the geometric structure of the transition $\mathbf{A}(x_t)$;
(ii) the architectural \textbf{family}; and (iii) explicit
\textbf{violations} of the canonical form, when present.

\paragraph{Choice of main-experiment models.}
The architectures evaluated in the main experiments
(\Cref{tab:canonical_comparison}, highlighted rows) span the canonical
form's $(\mathbf{A}, g, \phi)$ axes jointly: diagonal contractive (Mamba),
signed diagonal (Negative Mamba), damped complex rotation (Mamba-3),
unitary (AUSSM, Simple AUSSM), dense linear (Linear RNN), dense
pointwise-nonlinear (tanh RNN), input-gated (Token-gated RNN), and
state-gated (State-gated RNN), together with four delta-rule models with matrix-valued state (Gated DeltaNet, KDA, Gated DeltaNet-2, Gated DeltaProduct). Each realizes the canonical form, except for the violations documented below (the exponential-trapezoidal input injection of Mamba-3, and the matrix-valued state of the delta-rule models).
Architectures outside the main experiments are listed in
\Cref{tab:canonical_comparison} for reference: DeltaNet and DeltaProduct (reported in \Cref{subsec:ax_more_models})
violate the canonical form via matrix-valued state, while S4 is the
trivial input-independent case (a valid but degenerate canonical-form
realization).

\begin{sidewaystable}
\centering
\footnotesize
\begin{tabular}{llccccc}
\toprule
\textbf{Type} & \textbf{Model} & $g(h_{t-1}, x_t)$ & $\phi(\cdot)$ & $\mathbf{A}(x_t)$ & $b(x_t)$ & $\text{dec}(h_t, x_t)$ \\
\midrule

\multirow{8}{*}{SSM}
& S4~\citep{gu2022efficiently}
  & $1$
  & $\text{id}$
  & $(\mathbf{I} - \Delta/2 \cdot \mathbf{A})^{-1}$
  & $(\mathbf{I} - \Delta/2 \cdot \mathbf{A})^{-1}$
  & $\mathbf{C}h_t + \mathbf{D}x_t$ \\
&
  &
  &
  & $\times(\mathbf{I} + \Delta/2 \cdot \mathbf{A})$
  & $\times \Delta \mathbf{B} x_t$
  & \\
& \cellcolor{gray!15}Mamba~\citep{gu2024mamba}
  & \cellcolor{gray!15}$1$
  & \cellcolor{gray!15}$\text{id}$
  & \cellcolor{gray!15}$e^{\Delta_t \mathbf{A}}$
  & \cellcolor{gray!15}$\Delta_t \mathbf{B}_t x_t$
  & \cellcolor{gray!15}$\mathbf{C}_t h_t + \mathbf{D}x_t$ \\
& \cellcolor{gray!15}Negative Mamba~\citep{grazzi2024unlocking}
  & \cellcolor{gray!15}$1$
  & \cellcolor{gray!15}$\text{id}$
  & \cellcolor{gray!15}$2e^{\Delta_t \mathbf{A}} - \mathbf{I}$
  & \cellcolor{gray!15}$\Delta_t \mathbf{B}_t x_t$
  & \cellcolor{gray!15}$\mathbf{C}_t h_t + \mathbf{D}x_t$ \\
& \cellcolor{gray!15}Mamba-3~\citep{lahoti2026mamba3}
  & \cellcolor{gray!15}$1$
  & \cellcolor{gray!15}$\text{id}$
  & \cellcolor{gray!15}$e^{\Delta_t \mathbf{A}}$
  & \cellcolor{gray!15}$(1-\lambda_t)\Delta_t e^{\Delta_t \mathbf{A}}\mathbf{B}_{t-1}x_{t-1}$
  & \cellcolor{gray!15}$\mathbf{C}_t h_t + \mathbf{D}x_t$ \\
& \cellcolor{gray!15}
  & \cellcolor{gray!15}
  & \cellcolor{gray!15}
  & \cellcolor{gray!15}
  & \cellcolor{gray!15}$+ \lambda_t \Delta_t \mathbf{B}_t x_t$
  & \cellcolor{gray!15} \\
& \cellcolor{gray!15}AUSSM~\citep{karuvally2025aussm}
  & \cellcolor{gray!15}$1$
  & \cellcolor{gray!15}$\text{id}$
  & \cellcolor{gray!15}$e^{\Delta_t \mathbf{A}_t}$
  & \cellcolor{gray!15}$\Delta_t \mathbf{B} x_t$
  & \cellcolor{gray!15}$\mathbf{C} h_t + \mathbf{D}x_t$ \\
& \cellcolor{gray!15}Simple AUSSM~\citep{shakerinava2026expressive}
  & \cellcolor{gray!15}$1$
  & \cellcolor{gray!15}$\text{id}$
  & \cellcolor{gray!15}$e^{\mathbf{A}_t}$
  & \cellcolor{gray!15}$\mathbf{B} x_t$
  & \cellcolor{gray!15}$\mathbf{C} h_t + \mathbf{D}x_t$ \\

\midrule

\multirow{13}{*}{RNN}
& \cellcolor{gray!15}$\tanh$ RNN~\citep{elman1990finding}
  & \cellcolor{gray!15}$1$
  & \cellcolor{gray!15}$\tanh$
  & \cellcolor{gray!15}$\mathbf{W}_h$
  & \cellcolor{gray!15}$\mathbf{W}_x x_t + b_h$
  & \cellcolor{gray!15}$h_t$ \\
& \cellcolor{gray!15}Linear RNN
  & \cellcolor{gray!15}$1$
  & \cellcolor{gray!15}$\text{id}$
  & \cellcolor{gray!15}$\mathbf{W}_h$
  & \cellcolor{gray!15}$\mathbf{W}_x x_t + b_h$
  & \cellcolor{gray!15}$h_t$ \\
& \cellcolor{gray!15}Token-gated RNN
  & \cellcolor{gray!15}$\sigma(\mathbf{W}_g x_t + b_g)$
  & \cellcolor{gray!15}$\text{id}$
  & \cellcolor{gray!15}$\mathbf{W}_h$
  & \cellcolor{gray!15}$\mathbf{W}_x x_t + b_h$
  & \cellcolor{gray!15}$h_t$ \\
& \cellcolor{gray!15}State-gated RNN
  & \cellcolor{gray!15}$\sigma(\mathbf{W}_g x_t + \mathbf{U}_g h_{t-1}$
  & \cellcolor{gray!15}$\text{id}$
  & \cellcolor{gray!15}$\mathbf{W}_h$
  & \cellcolor{gray!15}$\mathbf{W}_x x_t + b_h$
  & \cellcolor{gray!15}$h_t$ \\
& \cellcolor{gray!15}
  & \cellcolor{gray!15}$+ b_g)$
  & \cellcolor{gray!15}
  & \cellcolor{gray!15}
  & \cellcolor{gray!15}
  & \cellcolor{gray!15} \\
& DeltaNet~\citep{yang2024deltanet}
  & $1$
  & $\text{id}$
  & $\mathbf{I} - \beta_t k_t k_t^\top$
  & $\beta_t k_t v_t^\top$
  & $h_t^\top q_t$ \\
& DeltaProduct~\citep{siems2025deltaproduct}
  & $1$
  & $\text{id}$
  & $\prod_{j=n_h}^{1} (\mathbf{I} - \beta_{t,j} k_{t,j} k_{t,j}^\top)$
  & $\sum_{j=1}^{n_h} \left( \prod_{k=n_h}^{j+1} (\mathbf{I} - \beta_{t,k} k_{t,k} k_{t,k}^\top) \right)$
  & $h_t^\top q_t$ \\
&
  &
  &
  &
  & $\times \beta_{t,j} k_{t,j} v_{t,j}^\top$
  & \\
& \cellcolor{gray!15}Gated DeltaNet~\citep{yang2025gated}
  & \cellcolor{gray!15}$1$
  & \cellcolor{gray!15}$\text{id}$
  & \cellcolor{gray!15}$\alpha_t(\mathbf{I} - \beta_t k_t k_t^\top)$
  & \cellcolor{gray!15}$\beta_t k_t v_t^\top$
  & \cellcolor{gray!15}$h_t^\top q_t$ \\
& \cellcolor{gray!15}KDA~\citep{kimiteam2025kimilinear}
  & \cellcolor{gray!15}$1$
  & \cellcolor{gray!15}$\text{id}$
  & \cellcolor{gray!15}$(\mathbf{I} - \beta_t k_t k_t^\top)\,\mathrm{Diag}(\alpha_t)$
  & \cellcolor{gray!15}$\beta_t k_t v_t^\top$
  & \cellcolor{gray!15}$h_t^\top q_t$ \\
& \cellcolor{gray!15}Gated DeltaNet-2~\citep{hatamizadeh2026gateddeltanet2}
  & \cellcolor{gray!15}$1$
  & \cellcolor{gray!15}$\text{id}$
  & \cellcolor{gray!15}$(\mathbf{I} - k_t (\beta^{\mathrm{e}}_t \odot k_t)^\top)\,\mathrm{Diag}(\alpha_t)$
  & \cellcolor{gray!15}$k_t (\beta^{\mathrm{w}}_t \odot v_t)^\top$
  & \cellcolor{gray!15}$h_t^\top q_t$ \\
& \cellcolor{gray!15}Gated DeltaProduct~\citep{siems2025deltaproduct}
  & \cellcolor{gray!15}$1$
  & \cellcolor{gray!15}$\text{id}$
  & \cellcolor{gray!15}$\alpha_t \prod_{j=n_h}^{1} (\mathbf{I} - \beta_{t,j} k_{t,j} k_{t,j}^\top)$
  & \cellcolor{gray!15}$\sum_{j=1}^{n_h} \left( \prod_{k=n_h}^{j+1} (\mathbf{I} - \beta_{t,k} k_{t,k} k_{t,k}^\top) \right)$
  & \cellcolor{gray!15}$h_t^\top q_t$ \\
& \cellcolor{gray!15}
  & \cellcolor{gray!15}
  & \cellcolor{gray!15}
  & \cellcolor{gray!15}
  & \cellcolor{gray!15}$\times \beta_{t,j} k_{t,j} v_{t,j}^\top$
  & \cellcolor{gray!15} \\
\bottomrule
\end{tabular}
\vspace{0.5em}
\caption{
\textbf{Recursive models mapped to the canonical form.} 
Comparison of various recurrent architectures mapped to our canonical form $h_t = \phi(g_t \odot (\mathbf{A}_t h_{t-1}) + b_t)$.
Mamba-3 utilizes an Exponential-Trapezoidal rule where the $b(x_t, x_{t-1})$ depends on both current and previous inputs. To maintain structural consistency, matrix-valued models such as DeltaNet \citep{yang2024deltanet} and DeltaProduct \citep{siems2025deltaproduct} are represented via a transpose transformation $h_t = \mathbf{S}_t^\top$, converting their original right-multiplication state updates into left-multiplication.
}
\label{tab:canonical_comparison}
\end{sidewaystable}

\subsection{State-\texorpdfstring{space models}{space models}}

\paragraph{S4~\citep{gu2022efficiently}.}
\textbf{Canonical form.} Constant transition:
$\mathbf{A}(x_t) = (\mathbf{I} - \Delta/2\,\mathbf{A})^{-1}(\mathbf{I} + \Delta/2\,\mathbf{A})$,
time- and input-invariant; $\mathbf{A}$ is HiPPO-initialized in
Normal-Plus-Low-Rank form for stable long-range dependence.
\textbf{Family.} Linear time-invariant structured state-space model.

\paragraph{Mamba~\citep{gu2024mamba}.}
\textbf{Canonical form.} Diagonal contractive:
$\mathbf{A}(x_t) = \exp(\Delta_t \mathbf{A})$ with real-valued diagonal
$\mathbf{A}\prec 0$ and $\Delta_t = \mathrm{softplus}(\Delta_{\mathrm{bias}} + W_\Delta x_t)$;
each entry of $\mathbf{A}(x_t)$ lies in $(0, 1)$.
\textbf{Family.} Diagonal selective state-space model.

\paragraph{Negative Mamba~\citep{grazzi2024unlocking}.}
\textbf{Canonical form.} Signed diagonal:
$\mathbf{A}(x_t) = 2\exp(\Delta_t \mathbf{A}) - \mathbf{I}$, sharing the
same diagonal $\mathbf{A}\prec 0$ as Mamba; entries of $\mathbf{A}(x_t)$
lie in $(-1, 1)$.
\textbf{Family.} Diagonal selective state-space model with signed
transitions.

\paragraph{Mamba-3~\citep{lahoti2026mamba3}.}
\textbf{Canonical form.} Damped complex rotation:
$\mathbf{A}(x_t) = \exp(\Delta_t \mathbf{A}_t)$ with complex-valued
diagonal $\mathbf{A}_t = \mathbf{A}_{\mathrm{re}}(x_t) + i\boldsymbol{\Theta}(x_t)$
(real decay $\mathbf{A}_{\mathrm{re}}(x_t)\prec 0$ and input-dependent
rotation frequencies $\boldsymbol{\Theta}(x_t)$); entries of
$\mathbf{A}(x_t)$ lie strictly inside the unit disk.
\textbf{Family.} Complex-diagonal selective state-space model.
\textbf{Violations.} Exponential-trapezoidal discretization makes the
input injection $b$ a linear combination of $\mathbf{B}_{t-1} x_{t-1}$
and $\mathbf{B}_t x_t$, whereas canonical $b$ takes $x_t$ only.

\paragraph{AUSSM~\citep{karuvally2025aussm}.}
\textbf{Canonical form.} Unitary:
$\mathbf{A}(x_t) = \exp(\Delta_t \mathbf{A}_t)$ with real
skew-symmetric input-dependent $\mathbf{A}_t$, so $\mathbf{A}(x_t)$ is
orthogonal at every step (equivalently, its eigenvalues lie on the
complex unit circle).
\textbf{Family.} Adaptive unitary state-space model.

\paragraph{Simple AUSSM~\citep{shakerinava2026expressive}.}
\textbf{Canonical form.} Unitary, complex-diagonal:
$\mathbf{A}(x_t) = \exp(i\,\boldsymbol{\Lambda}(x_t))$ with input-dependent
real $\boldsymbol{\Lambda}(x_t)$, so the diagonal log is purely imaginary
and entries of $\mathbf{A}(x_t)$ have unit modulus. AUSSM's
input-dependent step size $\Delta_t$ is dropped, since it is not needed
for representing groups~\citep{shakerinava2026expressive}.
\textbf{Family.} Complex-diagonal unitary state-space model.

\subsection{Recurrent \texorpdfstring{networks}{networks}}

\paragraph{Linear RNN.}
\textbf{Canonical form.} Constant dense transition:
$\mathbf{A}(x_t) = \mathbf{W}_h$, $\phi = \mathrm{id}$, $g \equiv 1$.
\textbf{Family.} Linear recurrent network.

\paragraph{tanh RNN~\citep{elman1990finding}.}
\textbf{Canonical form.} Constant dense transition with elementwise
nonlinearity: $\mathbf{A}(x_t) = \mathbf{W}_h$, $\phi(z) = \tanh(z)$.
The activation makes the per-step Jacobian state-dependent through
$\tanh'$.
\textbf{Family.} Elman recurrent network.

\paragraph{Token-gated RNN.}
\textbf{Canonical form.} Constant dense transition with input-only gate:
$\mathbf{A}(x_t) = \mathbf{W}_h$,
$g(x_t) = \sigma(\mathbf{W}_g x_t + b_g)$, $\phi = \mathrm{id}$. The gate
is independent of $h_{t-1}$, so the update remains affine in $h_{t-1}$.
\textbf{Family.} Input-gated linear recurrent network.

\paragraph{State-gated RNN.}
\textbf{Canonical form.} Constant dense transition with state-and-input
gate: $\mathbf{A}(x_t) = \mathbf{W}_h$,
$g(h_{t-1}, x_t) = \sigma(\mathbf{W}_g x_t + \mathbf{U}_g h_{t-1} + b_g)$,
$\phi = \mathrm{id}$. State dependence in the gate makes the per-step
Jacobian state-dependent.
\textbf{Family.} State-gated recurrent network.

\paragraph{LSTM~\citep{hochreiter1997lstm}.}
\textbf{Canonical form.} The cell-state update
$c_t = f_t \odot c_{t-1} + i_t \odot \tanh(\mathbf{U}_c h_{t-1} + \mathbf{W}_c x_t + b_c)$
realizes the canonical form on $c_t$, with the forget gate
$f_t = \sigma(\mathbf{U}_f h_{t-1} + \mathbf{W}_f x_t + b_f)$ as the
state-and-input gate and identity transition ($\mathbf{A} = \mathbf{I}$).
\textbf{Family.} Long short-term memory.
\textbf{Violations.} (i) Two coupled state variables (cell state $c_t$
and hidden state $h_t$); the canonical form has a single state.
Consequently, the input injection
$i_t \odot \tanh(\mathbf{U}_c h_{t-1} + \mathbf{W}_c x_t + b_c)$ for
$c_t$ depends on $h_{t-1}$ (the other state variable), whereas canonical
$b(x_t)$ takes only $x_t$. (ii) Output gating
$h_t = o_t \odot \tanh(c_t)$ applies a second nonlinear transformation
after the canonical update, outside the
$\phi(g \odot \mathbf{A} h_{t-1} + b)$ template.

\paragraph{GRU~\citep{cho2014learning}.}
\textbf{Canonical form.} Convex-combination update (Cho's convention):
$h_t = z_t \odot h_{t-1}
   + (1 - z_t) \odot \tanh\!\bigl(\mathbf{U}(r_t \odot h_{t-1}) + \mathbf{W} x_t\bigr)$,
with update gate $z_t = \sigma(\mathbf{U}_z h_{t-1} + \mathbf{W}_z x_t)$
in the role of the canonical state-and-input gate against identity
transition ($\mathbf{A}=\mathbf{I}$).
\textbf{Family.} Gated recurrent unit.
\textbf{Violations.} (i) The candidate term
$(1-z_t)\odot\tanh(\cdot)$ plays the role of $b$ but depends on $h_{t-1}$
(via the inner $\mathbf{U}(r_t \odot h_{t-1})$), whereas canonical $b(x_t)$
takes only $x_t$. (ii) The reset gate
$r_t = \sigma(\mathbf{U}_r h_{t-1} + \mathbf{W}_r x_t)$ multiplies
$h_{t-1}$ \emph{inside} the candidate, before the transition matrix
$\mathbf{U}$; the canonical form admits gating only outside the
transition. (iii) The update has the form
$g \odot h_{t-1} + (1-g) \odot \tanh(\cdot)$ rather than
$\phi(g \odot \mathbf{A} h_{t-1} + b)$, with the nonlinearity $\tanh$
applied only to the candidate, not to the full update.

\paragraph{DeltaNet~\citep{yang2024deltanet}.}
\textbf{Canonical form.} Generalized rank-1 Householder (delta-rule update):
$\mathbf{A}(x_t) = \mathbf{I} - \beta_t k_t k_t^\top$, with sigmoid-gated
scalar $\beta_t = \sigma(W_\beta x_t) \in (0, 1)$. The factor reduces to
a true Householder reflection only at $\beta_t = 2/\|k_t\|^2$; in general
it is a learnable rank-1 perturbation of the identity.
\textbf{Family.} Linear-attention recurrent network.
\textbf{Violations.} The state is matrix-valued
($S_t \in \mathbb{R}^{d_v \times d_k}$); we represent it as $h_t = S_t^\top$
to match the canonical left-multiplication form.

\paragraph{DeltaProduct~\citep{siems2025deltaproduct}.}
\textbf{Canonical form.} Product of $n_h$ generalized Householders:
$\mathbf{A}(x_t) = \prod_{j=n_h}^{1}(\mathbf{I} - \beta_{t,j} k_{t,j} k_{t,j}^\top)$,
each factor a learnable rank-1 update with sigmoid-gated $\beta_{t,j}$.
With sufficient $n_h$ and $\beta_{t,j}$ in the Householder regime, the
product can represent any orthogonal matrix (Cartan--Dieudonn\'{e}); in
general it spans a wider set.
\textbf{Family.} Multi-step linear-attention recurrent network.
\textbf{Violations.} Same matrix-valued state as DeltaNet.

\paragraph{Gated DeltaNet~\citep{yang2025gated}.}
\textbf{Canonical form.} Decayed generalized Householder:
$\mathbf{A}(x_t) = \alpha_t(\mathbf{I} - \beta_t k_t k_t^\top)$, with a
scalar decay $\alpha_t \in (0, 1)$ per head and $\ell_2$-normalized keys.
We allow negative eigenvalues~\citep{grazzi2024unlocking}:
$\beta_t = 2\sigma(W_\beta x_t) \in (0, 2)$, so the factor has eigenvalue
$\alpha_t(1 - \beta_t) \in (-\alpha_t, \alpha_t)$ along $k_t$ and
$\alpha_t$ on its orthogonal complement.
\textbf{Family.} Gated linear-attention recurrent network.
\textbf{Violations.} Same matrix-valued state as DeltaNet.

\paragraph{KDA~\citep{kimiteam2025kimilinear}.}
\textbf{Canonical form.} Gated DeltaNet with a per-channel decay:
$\mathbf{A}(x_t) = (\mathbf{I} - \beta_t k_t k_t^\top)\,\mathrm{Diag}(\alpha_t)$,
with $\alpha_t \in (0, 1)^{d_k}$ and $\beta_t \in (0, 2)$ as above.
\textbf{Family.} Gated linear-attention recurrent network.
\textbf{Violations.} Same matrix-valued state as DeltaNet.

\paragraph{Gated DeltaNet-2~\citep{hatamizadeh2026gateddeltanet2}.}
\textbf{Canonical form.} KDA with separate erase and write gates:
$\mathbf{A}(x_t) = (\mathbf{I} - k_t (\beta^{\mathrm{e}}_t \odot k_t)^\top)\,\mathrm{Diag}(\alpha_t)$
and $b(x_t) = k_t (\beta^{\mathrm{w}}_t \odot v_t)^\top$, with a channel-wise
erase gate $\beta^{\mathrm{e}}_t \in (0, 2)^{d_k}$ and write gate
$\beta^{\mathrm{w}}_t \in (0, 1)^{d_v}$. Setting both gates to the same
scalar recovers KDA.
\textbf{Family.} Gated linear-attention recurrent network.
\textbf{Violations.} Same matrix-valued state as DeltaNet.

\paragraph{Gated DeltaProduct~\citep{siems2025deltaproduct}.}
\textbf{Canonical form.} DeltaProduct with a scalar decay:
$\mathbf{A}(x_t) = \alpha_t \prod_{j=n_h}^{1}(\mathbf{I} - \beta_{t,j} k_{t,j} k_{t,j}^\top)$,
with $n_h = 2$, $\alpha_t \in (0, 1)$, and $\beta_{t,j} \in (0, 2)$.
\textbf{Family.} Multi-step gated linear-attention recurrent network.
\textbf{Violations.} Same matrix-valued state as DeltaNet.

\end{document}